\documentclass[conference]{IEEEtran}
\usepackage[T1]{fontenc}

\IEEEoverridecommandlockouts
\usepackage{cite}
\usepackage{amsmath,amssymb,amsfonts}
\usepackage{graphicx}
\usepackage{textcomp}
\usepackage{xcolor}
\usepackage{booktabs}

\usepackage{algorithm}
\usepackage{algorithmicx}
\usepackage{algpseudocode}
\usepackage{hyperref}

\newfloat{algorithm}{t}{lop}
\algnewcommand\algorithmichyperparams{\textbf{Hyperparameters:}}
\algnewcommand\AHyperparams{\item[\algorithmichyperparams]}

\algnewcommand\algorithmicparams{\textbf{Parameters:}}
\algnewcommand\AParams{\item[\algorithmicparams]}

\algnewcommand\algorithmicinput{\textbf{Input:}}
\algnewcommand\algorithmicoutput{\textbf{Output:}}
\algnewcommand\Input{\item[\algorithmicinput]}
\algnewcommand\Output{\item[\algorithmicoutput]}
\algnewcommand\algorithmicforeach{\textbf{for each}}
\algdef{S}[FOR]{ForEach}[1]{\algorithmicforeach\ #1\ \algorithmicdo}

\def\BibTeX{{\rm B\kern-.05em{\sc i\kern-.025em b}\kern-.08em
    T\kern-.1667em\lower.7ex\hbox{E}\kern-.125emX}}
\begin{document}

\title{Open-world Reconstruction-based Classifier with Concept Drift Adaptation}

\author{\IEEEauthorblockN{Joanna Komorniczak}
\IEEEauthorblockA{\textit{Department of Systems and Computer Networks} \\
\textit{Wrocław University of Science and Technology}\\
Wrocław, Poland \\
ORCID: 0000-0002-1393-3622}}

\maketitle

\begin{abstract}
Data stream processing has become central to modern machine learning applications, with \textit{concept drifts} and \textit{novel class appearances} posing the primary challenges faced by recognition methods. Adapting to concept drifts while also being able to detect novel classes and distinguish groups of unknown observations can be seen as a step towards effective method employment in an open-world setting. This work proposes a classification method that can detect and learn the distributions of new, unknown classes, as well as adapt to concept drifts once the distribution of known classes shifts. The proposed method uses the autoencoders' reconstruction errors and the \textsc{kde} that models their distribution. The autoencoders incrementally adapt to the changing distributions of known classes via weight fine-tuning, resulting in continuous adjustment to evolving concepts. Conducted experiments used a diverse set of synthetic data streams, embeddings of a real-world static \textsc{cifar} dataset arranged into a stream to simulate concept drifts and the emergence of unknowns, as well as real-world datasets where novelty is observed. The methodology was compared with state-of-the-art approaches with three-criterion recognition quality assessment and resource consumption evaluation. The results indicate that the proposed method is competitive with reference approaches in stationary environments and performs best when concept drifts occur.
\end{abstract}


\section{Introduction}

\begin{figure}[!htb]
    \centering
    \includegraphics[width=\linewidth]{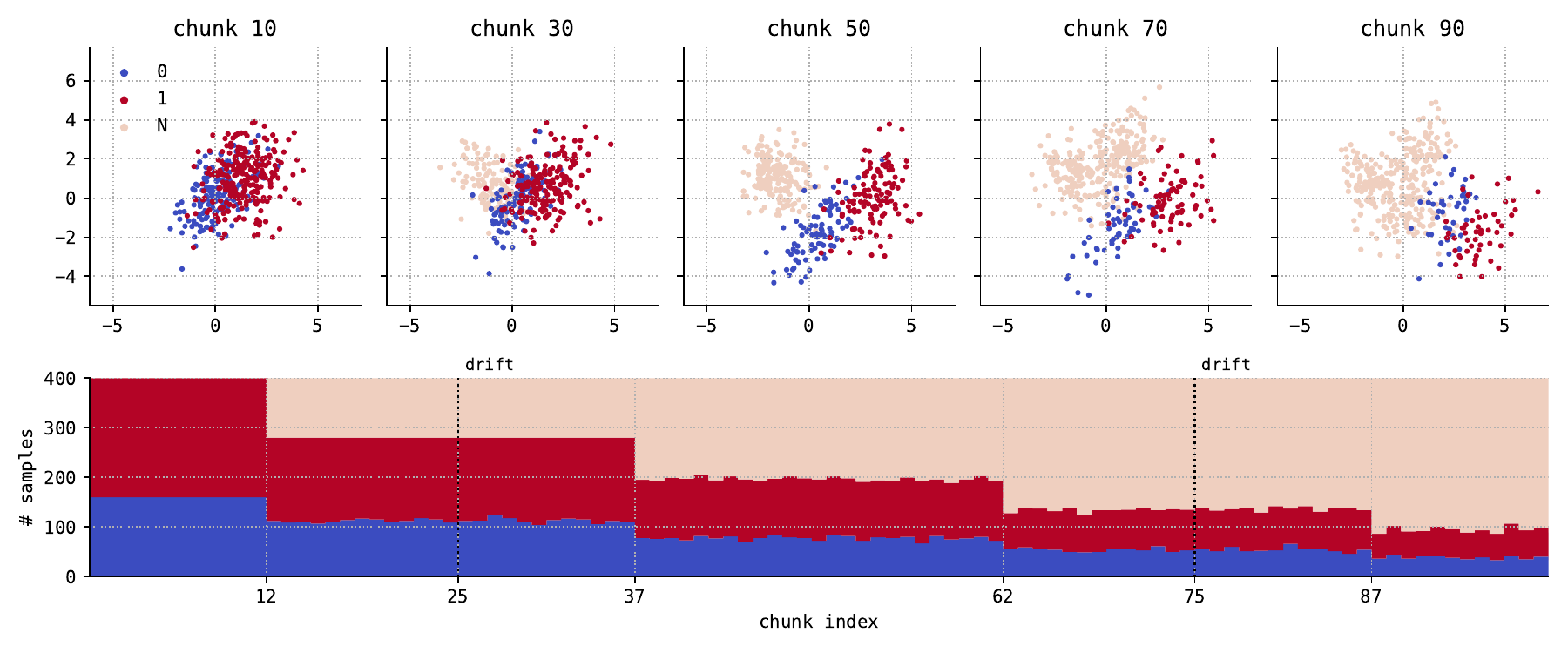}
    \caption{An example of a non-stationary data stream, where, over 100 chunks, four novel classes appear (beige markers). The known class distribution (blue and red markers) shifts due to concept drifts. The bottom plot shows the proportions of classes at a given timestamp.}
    \label{fig:vis_stream}
\end{figure}

Fueled by the growing popularity of digital media, today's world generates, processes, and stores immense amounts of data that align well with the paradigm of \textit{continuous} and \textit{dynamic} data stream processing \cite{de2016iot}. The modern machine learning systems employed for this purpose face various challenges, most often related to the non-stationarity of data, which may manifest as concept drift \cite{agrahari2022concept}. Since many systems process samples in real-time, the availability of labels for model training and adaptation may be limited \cite{gomes2022survey}. Moreover, some instances may represent previously unseen object categories -- a result of the emergence of novel, unknown observations \cite{gaudreault2024systematic}. In an \textit{open world} \cite{scheirer2012toward}, where the training set represents only part of reality, methods should be ready to signal shifts in sample distribution but also to mark instances that may represent previously unseen categories of objects. Figure \ref{fig:vis_stream} presents a data stream with the appearance of novel categories and a concept drift that affects the distribution of the known classes.

This article focuses on the complex task of processing non-stationary data streams, where both concept drift and unknown samples are present. The main contributions of this work are as follows:
\begin{itemize}
    \item The proposition of the \textsc{orcca} method for novelty detection. The method uses the autoencoder's reconstruction errors to classify known classes and recognize unknowns. The proposed method can distinguish groups of novelties among unknown instances and adapt to concept drifts without supervision via adjusted fine-tuning.
    \item A definition of a structured experimental environment that enables synthetic sampling of real data into a data stream with concept drifts among known classes and the emergence of unknowns. The experiments use 330 synthetic or semi-synthetic data streams and 3 large-scale real-world data streams.
    \item A three-criterion evaluation scheme that considers (a) recognition among known classes, (b) recognition of unknowns as a binary classification task, and  (c) recognition of various types of novelties. Evaluation considers the imbalanced nature of the novelty recognition task, where the proportion of known and unknown class samples changes over processing time.
    \item The analysis of the growth in processing time and memory usage of the proposed and reference methods.
\end{itemize}

The proposed methodology shifts the analysis from the entire set of features to just a one-dimensional \textit{proxy} of reconstruction errors. The method is evaluated on three real-world data streams as well as a large pool of synthetic and semi-synthetic data based on \textit{Madelon} distributions \cite{guyon2004result} and \textsc{cifar} embeddings. In semi-synthetic data streams, real class distributions are arranged in an order that simulates the arrival of unknowns. The known classes are additionally translated to simulate concept drifts. In real-world scenarios, the proposed solution can aid in network intrusion monitoring \cite{tavallaee2009detailed} or sensor-based activity tracking \cite{reiss2012introducing}, where reacting to evolving environments is essential. Unsupervised operation in an online phase makes the method independent of label costs and their potential delays.

\section{Related works}

The work connects to the tasks of concept drift adaptation, novelty detection, and open set recognition. This section presents the recent works related to those research areas in the data stream or incremental learning environment.

\paragraph{Concept drift detection}

Concept drift detection is an essential task in many systems operating in a data stream setting, as recognition methods may degrade due to a shift in data distribution \cite{gama2004learning}. Drift detectors aim to provide binary information on whether concept drift has occurred or the distribution remained stationary. The first generation of concept drift detectors directly employed the fact that the recognition quality of a classifier decreases as a result of a \textit{real} change in posterior probability $P(Y|X)$ distribution \cite{bifet2007learning}. More recent methods focus on analyzing the distribution of data -- allowing for implicit detection \cite{gozuaccik2021concept} -- and therefore for the recognition of \textit{virtual} drifts in covariate distribution $P(X)$ that do not yet lead to a loss of classifier performance. Among the implicit methods, unsupervised methods represent an important category of drift detection approaches \cite{lukats2024}. These methods are particularly valuable in applications where the arrival of labels may be delayed or their acquisition may be costly \cite{gomes2022survey}.

Implicit drift detectors can still utilize the classifier, which is done by the Margin Density Drift Detector (\textsc{md3}) \cite{sethi2015don}, but only to study the density of samples near the decision boundary. The Discriminative Drift Detector (\textsc{d3}) \cite{gozuaccik2019unsupervised} uses a logistic regression model to check if the distinction between the current and historical distributions represents an easy task. Other approaches, such as \textsc{ocdd} \cite{gozuaccik2021concept}, employ one-class classifiers to detect changes based on the number of reported outliers.

Deviations from the known data distributions can also be detected using autoencoders. One method that employs this idea is the Autoencoder Drift Detector (\textsc{add}) \cite{menon2020concept}, which was used to detect phishing attacks. Another case study used autoencoders for \textsc{eeg} analysis \cite{khadimallah2024autoencoder}, employing a separate model for each class in the data stream. 
\textsc{Aeddm} \cite{ali2025novel} uses two autoencoders that threshold their reconstruction errors to signal a drift in a binary classification setting. While most methods measure the reconstruction error cost function, some also rely on the cross-entropy \cite{jaworski2020concept}. 



The mentioned works that use autoencoders focus on the binary tasks of drift detection or anomaly detection. Meanwhile, an \textsc{orcca} proposed in this work is a complex multi-stage method that can classify among known classes, recognize unknowns and anomalies, learn novel class distributions, and adapt to concept drifts. Those aspects make the proposed method substantially different in terms of the scope of the task and the implemented mechanisms.

\paragraph{Novelty detection}

The emergence of a new category of objects -- even in an unsupervised setting, where the labels of the novel class have not yet arrived -- affects the data covariate probability $P(X)$. It can therefore be interpreted as a specific case of concept drift. In contrast to a typically considered concept drift, which affects only known classes, this type of change will not be recognized by methods that monitor the classification errors until the model is explicitly adapted to the evolving recognition problem. In contrast to the straightforward binary output of concept drift detectors (i.e. distribution remained static/ distribution changed), the purpose of novelty detectors is to mark the unknowns and group them into separate classes.

While autoencoders find applications in drift and outlier detection, most existing novelty detection methods directly analyze the neighborhood of original samples or the distances between clusters \cite{faria2016novelty}. One of the first methods for novelty detection, \textsc{olindda} \cite{spinosa2009novelty}, relies on clusters formed using the \textit{k-means} algorithm. This novelty detector is limited to a binary task of recognizing known instances and novelties, similar to one-class classifiers \cite{khan2009survey}. ECSMiner \cite{masud2010classification} employs an ensemble of classifiers to recognize among known classes. To detect novelty, it clusters the samples stored in short-term memory and analyzes the distances between known and unknown examples using a measure named \emph{q}-\textsc{nsc}. The method by default operates in a supervised manner and in batch-processing mode. In later research, it has been adapted to an unsupervised setting (\textit{without feedback} \cite{de2016minas}). ECSMiner can recognize among many known classes and detect novelties, but it requires labels for the concept drift adaptation mechanism. This limitation has been addressed in \textsc{echo} \cite{haque2016efficient}, which adapts with partial supervision, requesting the labels only for the samples with low prediction support. Both ECSMiner and \textsc{echo} react to concept change by updating the ensemble structure. Another method -- \textsc{minas} \cite{de2016minas}, similarly to \textsc{olindda}, relies on clustering. In contrast to the predecessor, \textsc{minas} can recognize novelties when multiple known classes are considered. \textsc{Minas}, like \textsc{olindda}, also operates in a fully unsupervised manner after an offline phase. A particular advantage of \textsc{minas} is its ability to recognize \textit{novelty patterns} when novelties are not grouped in a single cluster. The adaptation to concept drifts is linked to the continuous update of micro-clusters.

Those novelty detectors describe a class as a set of micro-clusters. To detect novelty, they rely on thresholding the distances between cluster centroids and incoming samples. In one interesting work where both concept drifts and novelties were detected \cite{shang2025novelty}, the authors used the \textit{radial base distance} to detect significant distribution changes. The work considers the benefits of dimensionality reduction and mentions \textit{principal component analysis} as a tool for that purpose.

The proposed \textsc{orcca} does not rely on raw distances in multi-dimensional space. The autoencoder learns the structure of a class and proves to be suitable for data where classes are not represented by compact groups of adjacent instances, e.g., in  high-dimensional embedding space. \textsc{Orcca}, similarly to \textsc{minas}, allows for discovering more than one class at a time but relies on reconstruction error distances instead of analyzing raw features. Similarly to \textsc{echo}, it can explicitly adapt to concept drifts, but it does so in a fully unsupervised manner.

\paragraph{Open set recognition}

When novelty detection shifts from tabular to unstructured data, many concepts overlap with \textit{open set recognition} (\textsc{osr}) \cite{gaudreault2024systematic}. The aim of the task is to correctly classify known class samples while being able to recognize unknown instances. The second objective is directly related to anomaly detection, with a major difference being that the unknown instances are meant to appear more frequently than the anomalies \cite{scheirer2012toward}.

Many \textsc{osr} methods have been proposed and evaluated for unstructured data like images and text. Methods often rely on distance calculations in the latent space or employ contrastive training to explicitly account for unknowns \cite{ge2017generative}. One of the methods that employ encoders is \textsc{c2ae} \cite{oza2019c2ae}, which trains the encoder and decoder in two separate stages to allow for classification among known classes. The method models the reconstruction errors using extreme value theory, often employed in \textsc{osr}. 

Even though unknown observations are a hallmark of a dynamic data stream environment, \textsc{osr} remains relatively rarely addressed in a streaming setting \cite{barcina2024resilience}. Most other propositions in this field are designed and evaluated on static and stationary data.




\section{Method}

This work proposes an \underline{O}pen-world \underline{R}econstruction-based \underline{C}lassifier with \underline{C}oncept Drift \underline{A}daptation (\textsc{orcca}) method.

The proposed algorithm combines the idea of using an ensemble of trained per-class autoencoders, reconstruction error thresholding, and density estimation to classify known samples and recognize unknowns. The relative reconstruction error $\epsilon$ of a sample $x$ is described in Equation (\ref{eq:rel-error}), where $\mathcal{A}$ denotes a trained autoencoder.

\begin{equation}
\epsilon(\mathcal{A}, x) = \frac{\|x - \mathcal{A}(x)\|_1}{\|x\|_1}
\label{eq:rel-error}
\end{equation}

The relative reconstruction error is calculated using the L1 norm. The sum of absolute reconstruction errors is divided by the sum of absolute feature values of the whole sample ($\|x\|_1$).

The unknowns are grouped into new classes based on the distribution of their relative reconstruction errors and the responses of kernel density estimation (\textsc{kde}) $\mathcal{K}$, fitted on the scaled reconstruction errors $E = [e_1, ..., e_n]$, where $e(x) = \frac{\epsilon(\mathcal{A}, x)}{\sigma}$ and $\sigma$ denotes a standard deviation of validation set errors in a considered class, established during ensemble extension. $K$ is the Gaussian kernel, and $h$ follows Scott's rule. The model is used to establish density for scaled error $e$, as described in Equation~(\ref{eq:kde}).

\begin{equation}
    \mathcal{K}(e) = \frac{1}{n h} \sum_{i=1}^{n} K\left(\frac{e - e_i}{h}\right)
    \label{eq:kde}
\end{equation}

The method uses one-sided estimation $\mathcal{K}'$. When estimating the density for a given model $c$, an error $e_c$ below a mean value of this class errors $\mu_c$ is clipped to the mean, as described in Equation (\ref{eq:os-kde}). Reconstruction that is better than average for this class will not lower the estimation.
\begin{equation}
    \mathcal{K}'_c(e_c) = \mathcal{K}_c\big(\max(e_c, \mu_c)\big)
    \label{eq:os-kde}
\end{equation}
    
\textsc{Orcca} also employs autoencoder fine-tuning to allow adaptation to non-stationary concepts. Using the reconstruction errors instead of analyzing the original features or their embeddings in the high-dimensional space makes the method suitable for the time and memory constraints of data stream processing systems. Fine-tuning is performed with a given global probability $p$. The adjusted probability $\pi = min(1, 2pg)$ is calculated according to the fraction of samples $g$ whose error was greater than the current one in the recent time window of $w$ samples. If the error is small, the probability $\pi$ increases. When past errors for a model hold fewer than 20 values, $\pi = p$.

\begin{figure}[!htb]
    \centering
    \includegraphics[width=\linewidth]{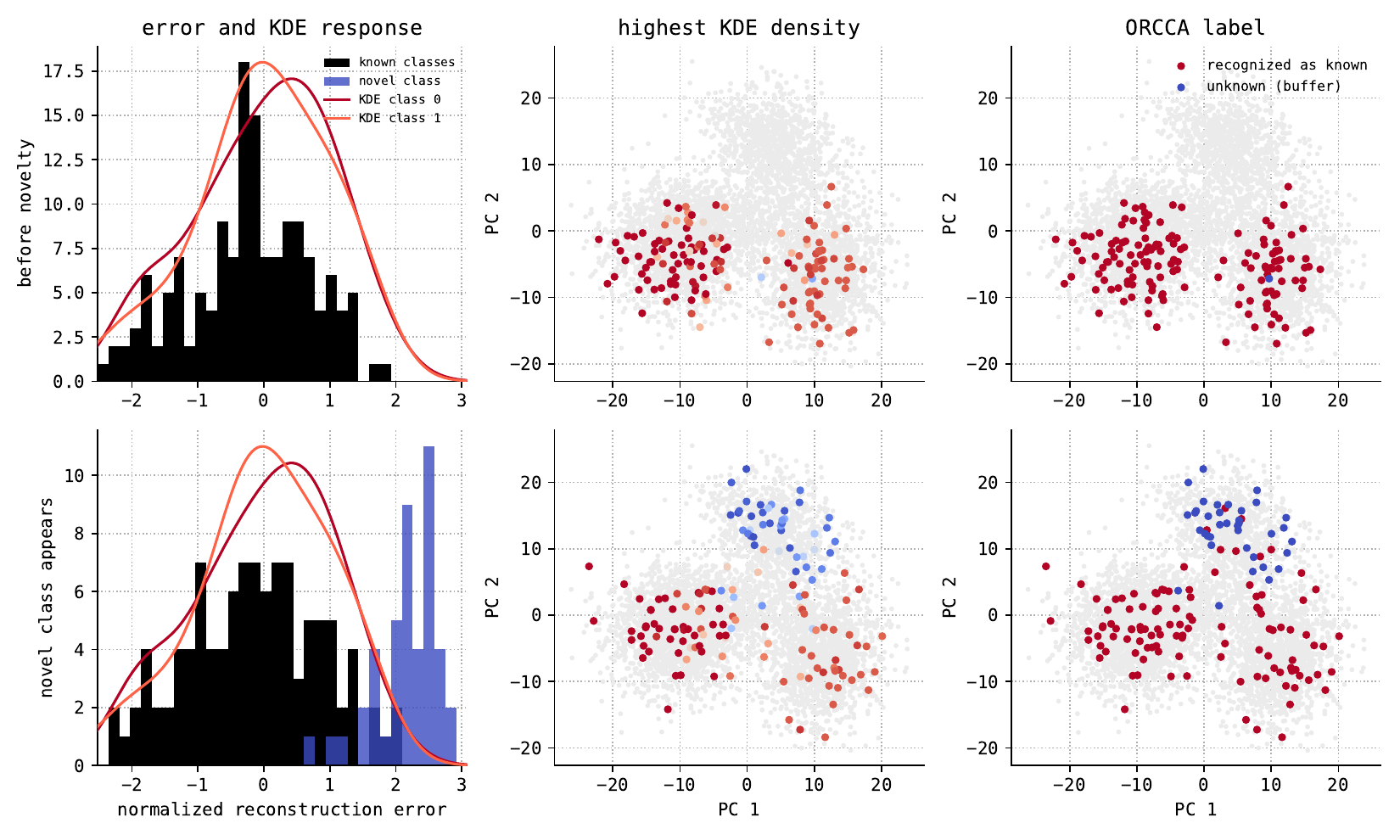}
    \caption{The normalized distributions of reconstruction errors of known classes (black bars) and novel-class samples (blue bars). The central column shows the responses of KDE (red color indicates high density score) and the right column the label known/unknown produced by \textsc{orcca}. Scatterplots use two principal components. }
    \label{fig:example_kde}
\end{figure}

The visual presentation of the method's main operation principle when faced with novel class appearance is shown in Figure \ref{fig:example_kde}. The images show two stream windows: first, when only known classes are present, and, in the second row of the figure, when the new class arrives. The first column shows the distributions of errors and the \textsc{kde} density of known classes (black bars and red lines). The error distribution of new class instances, shown in blue, clearly differs from the known-class reconstruction errors. Scatterplots show the principal components of the stream. The color of the points in the central column indicates the density based on \textsc{kde}, while the color in the right column indicates the predicted label in a binary known vs unknown recognition problem.

\begin{figure}[!htb]
    \centering
    \includegraphics[width=\linewidth]{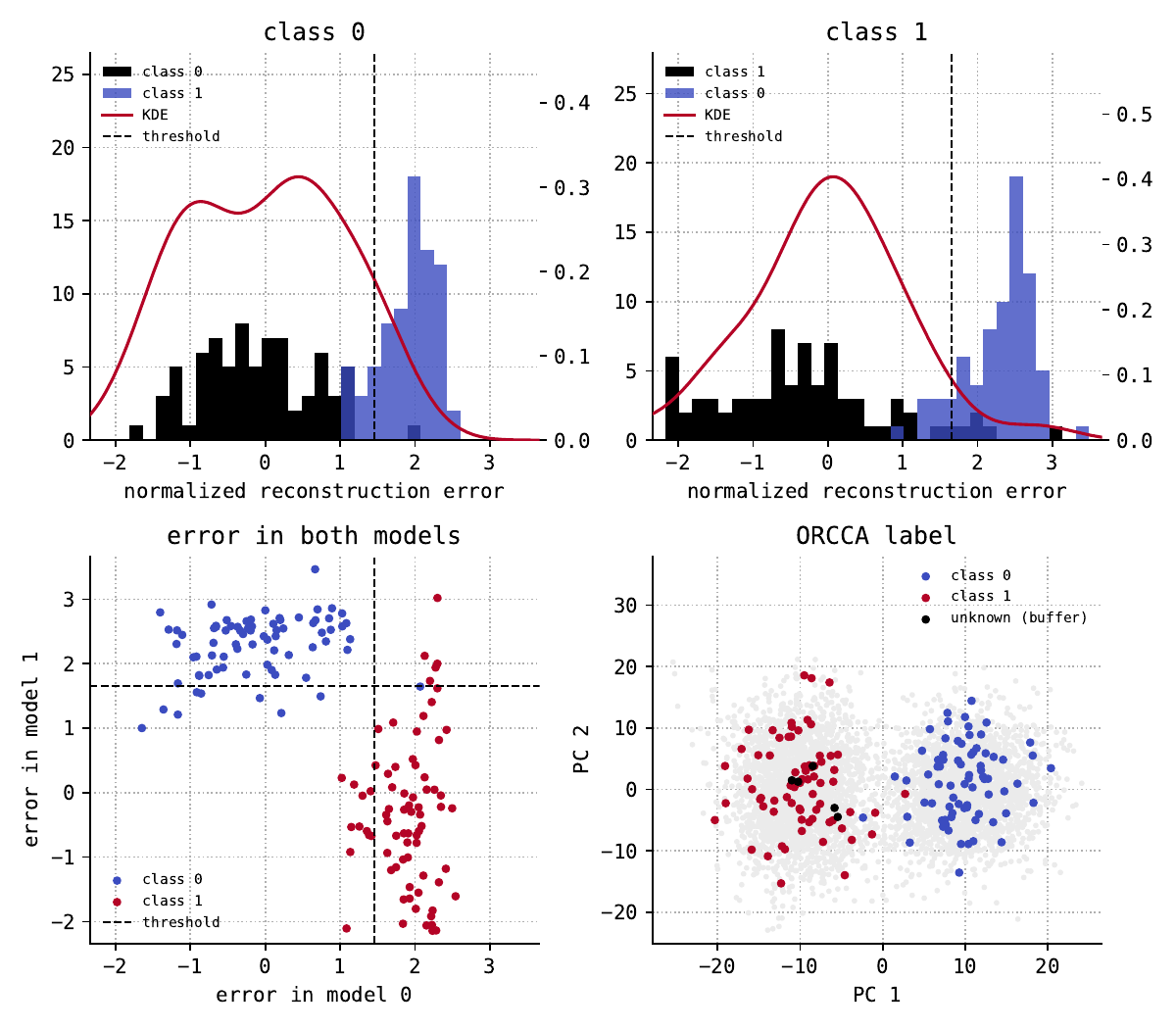}
    \caption{The error distributions of two known classes when samples are reconstructed by both autoencoders. The black bars indicate errors of a given class, and blue bars -- errors of the other class. The scatterplots show the distribution of errors with thresholds (dashed lines) and the principal components with labels on the last plot. The three samples that fell outside of both thresholds were marked as unknowns (black points).}
    \label{fig:example_knowns}
\end{figure}

Additionally, Figure \ref{fig:example_knowns} shows the operation of the thresholds employed in \textsc{orcca} that are used when classifying known instances. Those thresholds are established during the offline phase by seeking the point between the reconstruction errors of a given class (validation subset) and the remaining samples. The use of per-class thresholds aims to increase the recognition quality among known classes, whose labels are explicitly provided.

The threshold $t_c$ for model $c$ is calculated based on the errors of the validation set $X_{val}$, isolated from the provided samples $X_{c}$ of class $c$. The autoencoder $\mathcal{A}_{c}$ provides errors of $X_{val}$ (denoted as $E$) and samples of the other known classes $X_{neg}$ (denoted as $E^-$). Threshold $t_c$ lies at an equal distance from both error distribution means, as shown in Equation (\ref{eq:threshold}), where $\mu$ and $\sigma$ are the mean value and standard deviation, respectively.
\begin{equation}
    t_c = \frac{\mu(E)\,\sigma(E^-) + \mu(E^-)}{1 + \sigma(E^-)}
    \label{eq:threshold}
\end{equation}

The threshold $t_c$ and \textsc{kde} model $\mathcal{K}_c$ are established only once, when the ensemble is extended with models for class $c$. Adaptation is applied only to known-class models and aims to bring the reconstruction errors of a drifting class back to the calibrated distribution, so the fixed $t_c$ and $\mathcal{K}_c$ density serve as reliable references.

A detailed description of the method is presented in Pseudocode \ref{alg:orca}. The algorithm accepts a data stream $\mathcal{DS}$ with the labels provided during an offline phase. In experiments, the offline phase consisted of 250 to 2000 samples. As a hyperparameter, one can specify the architecture of a fully connected autoencoder $\mathcal{A}$ (e.g., the number of hidden layers, bottleneck width), the size of a buffer $B$, the proportion $\rho$ of the given class isolated for validation, \textsc{kde} threshold $\tau$, the bimodality threshold $\beta$, the adaptation probability $p$, and window for weighted adaptation $w$. The known classes will be labeled according to the labels provided in the offline phase, and the recognized unknown classes will receive the following indices. The method was designed for a short offline phase to enable immediate deployment without the need to thoroughly collect samples for training. For this reason, the offline phase length is equivalent to a buffer size $B$. The buffer $\mathcal{B}$ that is used to store samples in the offline phase is later reused as novelty buffer.

\begin{algorithm}[!htb]
\scriptsize
\caption{The pseudocode of the \textsc{orcca} method}
\label{alg:orca}
\begin{algorithmic}[1]
\Input
\Statex $\mathcal{DS} = [x_1, x_2, \ldots ]$ -- data stream, labels $y_t$ for the first $B$ samples

\AHyperparams
\Statex $\mathcal{A}$ -- fully connected autoencoder,
$B$ -- buffer size,
$\rho$ -- validation fraction,
$\tau$ -- novelty threshold on KDE density,
$\beta$ -- bimodality threshold,
$p$ -- adaptation probability,
$w$ -- adaptation window length

\AParams
\Statex $\mathcal{M} = [M_1, M_2, \ldots]$ -- ensemble, $M_m = (\mathcal{A}_m, \mathcal{K}_m, t_m)$: autoencoder, kernel density estimation model, threshold,
$\mathcal{B}$ -- buffer of samples,
$\mathcal{E}$ -- buffer of reconstruction errors

\Output
\Statex $[\hat{y}_1, \hat{y}_2, \ldots ]$ -- class indices $\hat{y}_t$
\vspace{1em}

\State $\mathcal{M} \gets \emptyset$, $\mathcal{B} \gets \emptyset$, $\mathcal{E} \gets \emptyset$

\ForAll{$x_t \in \mathcal{DS}$}

    \If{$t \leq B$}
    {\color{gray}\Comment{offline phase}}
        \State $\mathcal{B} \gets \mathcal{B} \cup \{(x_t, y_t)\}$
        \If{$|\mathcal{B}| = B$}
            \ForAll{class $c$ in $\mathcal{B}$}
            {\color{gray}\Comment{one model per known class}}
                \State $X_c \gets$ samples of class $c$ in $\mathcal{B}$
                \State $X_{tr}, X_{val} \gets$ split $X_c$ with $\rho$                
                \State $\mathcal{A}_c \gets$ copy of $\mathcal{A}$ trained on $X_{tr}$
                \State $\mathcal{K}_c \gets \mathcal{K}$ fitted on scaled errors of $X_{val}$
                \State estimate $t_c$
                {\color{gray}\Comment{incrementally build the ensemble}}
                \State $\mathcal{M} \gets \mathcal{M} \cup \{(\mathcal{A}_c, \mathcal{K}_c, t_c)\}$
                                
            \EndFor
            \State $\mathcal{B} \gets \emptyset$
        \EndIf
        \State \Return
    \EndIf
    {\color{gray}\Comment{online phase}}

    \State $C \gets$ known classes whose $e(x_t)$ is below their threshold
    \If{$C \neq \emptyset$}
        \State $\hat{y}_t \gets$ class in $C$ with the lowest $e(x_t)$
        {\color{gray}\Comment{best fitting known class}}
        \State fine-tune $\mathcal{A}_{\hat{y}_t}$ with $x_t$ and probability $\pi$
        {\color{gray}\Comment{adapt to concept drift}}
        \State \Return $\hat{y}_t$
    \EndIf

    \ForAll{novel-class models $M_m \in \mathcal{M}$, from the oldest}
        \If{$\mathcal{K}'_m$ finds $e(x_t)$ typical}
            \State \Return $m$
            {\color{gray}\Comment{first class in cascade}}
        \EndIf
    \EndFor
    {\color{gray}\Comment{no model recognized sample}}
    \State $\mathcal{B} \gets \mathcal{B} \cup \{x_t\}$
    \State $\mathcal{E} \gets \mathcal{E} \cup \{ e(x_t)$ in all models$\}$
    \If{$|\mathcal{B}| = B$}
        \State \Call{ExtendEnsemble}{$\mathcal{B}$, $\mathcal{E}$}
        {\color{gray}\Comment{learn a new distribution}}
    \EndIf
    \State \Return $|\mathcal{M}| + 1$
    {\color{gray}\Comment{label as the novel class}}

\EndFor
\end{algorithmic}

\end{algorithm}

In the offline phase (lines 3:17), the data is split into separate classes and then into training and validation sets $X_{tr}$, $X_{val}$. An autoencoder $\mathcal{A}_c$ for class $c$ is trained with  $X_{tr}$. The isolated validation set $X_{val}$ is used to measure reconstruction errors, and those errors, after scaling with $\sigma$, are used to build a \textsc{kde} $\mathcal{K}_c$. In this stage, the threshold $t_c$ is also established. The buffer $\mathcal{B}$ that was used to collect offline samples is reset and will be reused for novelties.

In the online phase, the method first considers known classes (lines 18:23). $C$ is a set of candidate classes whose scaled error $e$ falls below their threshold. If a sample is recognized as any of the known classes, the label $\hat{y}_t$ is set as the class with the lowest error. Next, the autoencoder $\mathcal{A}_{\hat{y_t}}$ is fine-tuned to allow adaptation to concept drifts. In cases where none of the known-class models recognize the sample, the novel-class models are considered (lines 24:28). The cascade of \textsc{kde} estimation is performed, starting from the oldest model. When the one-sided density of model $c$ is above the density threshold of \textsc{kde} ($\tau$), the label $\hat{y}_t$ is set to class $c$. The cascade intends to reduce the processing time when querying autoencoders.

The last section of the pseudocode describes the procedure when none of the known and unknown-class models recognize the sample (lines 29:34). The sample is added to the buffer, as well as their errors in each autoencoder. The sample is labeled as the novel class $\hat{y}_t = |\mathcal{M}| + 1$. When the buffer is full, the steps described in Algorithm \ref{alg:alg-extend} are performed.

\begin{algorithm}[!htb]
\scriptsize
\caption{\textsc{ExtendEnsemble} function, performed when a model for novelty is added}
\label{alg:alg-extend}
\begin{algorithmic}[1]
\Input
\Statex $\mathcal{B}$ -- full buffer of novel samples,
$\mathcal{E}$ -- their reconstruction errors in all models

\vspace{1em}

    \State $b \gets$ BC of the errors of each model in $\mathcal{E}$
    \If{$\max(b) \leq \beta$}
        \State $X \gets \mathcal{B}$
        {\color{gray}\Comment{errors unimodal -- one new class}}
    \Else
        \State fit a two-component GMM on $\mathcal{E}_m \{m \in \mathcal{M} : b_m > \beta \}$
        \State $X \gets$ samples of the more numerous modality
        {\color{gray}\Comment{learn the dominant modality}}
    \EndIf

    \State $X_{tr}, X_{val} \gets$ split $X$ with $\rho$
    \State train $\mathcal{A}_c$ on $X_{tr}$
    {\color{gray}\Comment{warm start from the previous model}}
    \State fit $\mathcal{K}_c$ on scaled errors of $X_{val}$
    \State $\mathcal{M} \gets \mathcal{M} \cup \{(\text{copy of } \mathcal{A}_c, \mathcal{K}_c)\}$
    \State remove $X$ and samples recognized by $\mathcal{A}_c$ from $\mathcal{B}$
    \State $\mathcal{E} \gets$ errors of the remaining samples of $\mathcal{B}$
\end{algorithmic}

\end{algorithm}

The \textsc{orcca} method can distinguish between more than one novel class at a time. The grouping of unknown samples, stored in a buffer, is performed according to the errors in $\mathcal{E}$. The procedure calculates Sarle's bimodality coefficient (\textsc{bc}) $b_j$ for errors with skewness $\gamma_j$ and kurtosis $\kappa_j$ from each existing model $j$, as presented in Equation (\ref{eq:sarle}). 
\begin{equation}
    b_j = \frac{\gamma_j^2 + 1}{\kappa_j}
    \label{eq:sarle}
\end{equation}

If none of the error distributions exceed $\beta$, the buffer is assumed to be unimodal (line 3). A default $\beta=\frac{5}{9}$ value characterizes uniform distributions. Otherwise, the Gaussian Mixture Model (\textsc{gmm}) groups the errors that present a bimodal distribution. The more numerous group of objects separated by \textsc{gmm} is used to fit the new model (line 6). Training in this stage uses \textit{warm start} from previous novelty model. Using the entire buffer to fit a new autoencoder could lead to \textit{contamination} of $\mathcal{A}_c$ with outliers or samples of rare classes present in the buffer. After autoencoder training, samples of the dominant modality are removed from the buffer. Since both \textsc{bc} calculation and \textsc{gmm} grouping may be unreliable, an additional step is performed to check if samples of the less numerous modality are also correctly reconstructed by $\mathcal{A}_c$ (line 12). At the final step, the error buffer $\mathcal{E}$ is updated with the scaled errors of a new model.

Using the previous novelty model as a baseline autoencoder intends to help in weight optimization. The default recognition strategy in the case of novelty uses a cascade rule. The new model, adapted to both previous and new distributions, will still account for recognition towards a novel class.

Since \textsc{orcca} relies on fully connected autoencoders, the selection of its architecture will impact the results. For the presented study, an encoder and decoder with a single hidden layer containing 64 neurons were chosen. The bottleneck has 10 neurons. The architecture uses the ReLU activation function. By default, the method uses the Adam optimizer and a learning rate of $10^{-3}$ during the training, and $10^{-4}$ during adaptation. Initial training used 150 epochs and the \textit{mean squared error} loss function. In adaptation, a single optimization step is performed. The method returns predictions for each sample. By default, \textsc{orcca} uses $\rho=0.3$, $\beta=\frac{5}{9}$, $\tau=0.275$, $w=200$, and $p=0.6$. The default parameters of \textsc{orcca} were selected in preliminary experiments on synthetic tabular data streams with configurations disjoint from those used in the following evaluation.

\section{Experimental environment}

This section presents the set-up of experiments that compare \textsc{orcca} with reference approaches. The implementation of the method, the references, experiments, and their results are publicly available as a GitHub repository\footnote{\url{https://github.com/JKomorniczak/orcca}}. The implementation used Python with \textit{scikit-learn}, \textit{pytorch}, \textit{river} \cite{montiel2021river}, and \textit{streamndr} \cite{gaudreault2024systematic} libraries.

\subsection{Data streams with concept drift and class novelties}

The method was designed for data streams that exhibit both concept drift and class novelty. Besides using real-world data, similar to existing research in the field \cite{cejnek2018concept,masud2010classification}, experiments also use synthetic data where the number of events (concept drifts and novelties) and other data stream parameters could be configured. The research utilized 330 synthetic and semi-synthetic data streams generated with the \textsc{owdsg} generator \cite{komorniczak2025synthetic} based on fully-synthetic Madelon \cite{guyon2004result} datasets and using \textsc{cifar}-10 embeddings from a pre-trained ResNet18 model.

The generator used in this research was an extended version of the original method, publicly available in the \textsc{orcca} repository. The extension involved (a) sampling from existing distributions in order to generate \textsc{cifar}-based data streams, and (b) introducing smooth concept transitions by translating class distribution in a random direction. The second modification aimed to allow studying adaptation ability in an unsupervised online phase. The probability of each novel class occurrence in a given time window was set to 30\%. The data streams were processed sample-by-sample. Table~\ref{tab:generator} shows the complete configuration of the data stream generation procedure. The replications of each set-up enabled a stabilization analysis of the obtained results. In \textsc{cifar}-based streams the classes were shuffled randomly, each time presenting different known and unknown distributions, as well as a different novel class order.

\begin{table}[!htb]
    \caption{The configuration of synthetic / semi-synthetic data stream generation used in the experiments.}
    \centering
    \setlength{\tabcolsep}{5pt}
    \renewcommand{\arraystretch}{1.1}
    \scriptsize
    \begin{tabular}{l|ll|l}
    \toprule
         & \multicolumn{2}{c|}{\textsc{comparison}} & \textsc{resource use} \\
         \textsc{hyperparameters} & \textsc{tabular} & \textsc{cifar-10}  & \textsc{tabular} \\
         \midrule
         source distribution & Madelon & ResNet-18  & Madelon \\
         features (informative) & 100 (50) & 512  & 100 (50) \\
         samples & 5,000 & 5,000 & 100,000 \\ 
         known + unknown classes & 2+8, 4+6, 6+2 & 2+8, 4+6, 6+2  & 3+0, 3+3, 3+5 \\
         concept drifts & 0, 1, 3, 5, 7 & 0, 1, 3, 5, 7 & 0, 4 \\
         class separation & 2.5 & --  & 2.0 \\
         replications & 10 & 10  & 5 \\
    \bottomrule
    \end{tabular}
    \label{tab:generator}
\end{table}

Besides the generated data streams, three real-world data streams were used:
\begin{itemize}
    \item Covertype -- a data stream with 54 features and 581,012 samples. Three out of 7 classes were considered known. The least numerous class contains 0.47\% of the data, and the novelties comprise 13\% of the samples present in the stream. The natural concept drifts in this data stream come from the spatial location of forest coverage when collecting the data \cite{blackard1999comparative}.
    \item \textsc{kdd} Cup '99 (10\%) -- a data stream with 116 features and 494,019 samples. The features come from one-hot encoding of categorical attributes and the removal of invariant ones. Only one class was known, and 22 rare classes appeared during processing. The least numerous class contains only two samples, and in total, 80\% of the data stream comprises novelties that represent various attack types. The data describes extreme imbalance, where 20 new classes sum up to 1.7\% of all samples \cite{tavallaee2009detailed}.
    \item \textsc{pamap2} -- a data stream with 40 features and 194,292 samples, describing the sequence of activities performed by 9 people. One class was considered known, and 11 new classes appeared. The least numerous class represents 2.5\% of the samples, and in total, 90\% of the samples are novelties. The difficulty of this dataset may come from various characteristics of activities when performed by a different person \cite{reiss2012introducing}.
\end{itemize}

In the data streams, the original order of samples was maintained. This resulted in the presence of only one known class in two real-world data streams. In this case, \textsc{orcca} was unable to calibrate a threshold based on contrastive samples. The threshold was set according to the errors $E$ of the known class only as $t = \mu(E) + 2.5 \cdot \sigma(E)$. Such a strategy of establishing $t$ is the default in \textsc{orcca} when a single class is present in an offline phase.

\subsection{Reference methods}

The experiments focused on the ability of methods to (a) classify knowns, (b) recognize novel samples, and (c) group novelties into categories. The labels were delivered to the methods during the short offline training phase, and the methods operated in a fully unsupervised manner -- without any feedback -- in the online phase. This presumption made it necessary to adapt some methods to the unsupervised setting.

In this setting, the number of considered classes increased throughout the data stream processing. Experiments compared the proposed \textsc{orcca} with seven reference methods: three established as novelty classifiers, two stream clustering algorithms, and two canonical models employing one-class classification and confidence thresholding, adapted to the novelty detection task. The reference methods and their configurations are as follows:

\begin{itemize}
    \item \textsc{minas} \cite{de2016minas} with $k_{ini}=3$ micro-clusters, a threshold factor for extending existing classes is set to $1.1$ and a window size of $1000$. The minimum number of examples in a cluster is $10$. This configuration is the default one in a source implementation \cite{gaudreault2024systematic}.
    
    \item ECSMiner \cite{masud2010classification} without feedback, as implemented in \cite{de2016minas}. The method has $k=5$ micro-clusters per classifier and an ensemble size of $3$. The minimum number of examples in a cluster is $20$. The algorithm uses K-means clustering.

    \item \textsc{echo} \cite{haque2016efficient} with $3$ classifiers in the ensemble and $k=5$. The minimum number of examples in a cluster is $20$, and K-means clustering is also used. Window width for change detection is $W=500$. Due to an unsupervised setting in the online phase, the model's own predictions were used as feedback.
   
    \item CluStream \cite{aggarwal2003framework} -- a clustering algorithm employed in the novelty detection framework, with at most $100$ micro-clusters and a time window of $1000$ samples. The method is the only one that requires the true number of classes. The micro-cluster radius factor was adjusted to the feature scale.

    \item DenStream \cite{cao2006density} -- a clustering algorithm employed in the novelty detection framework, initialized with $\mu=3$, $\beta=0.5$, and a decay factor $\lambda=0.25$. The radius was adjusted to the feature scale.

    \item One-class Novelty Detector (\textsc{ocnd}) -- a method that uses a one-class \textsc{svm} classifier to recognize unknown class samples. The method uses the \textsc{svm} with an \textsc{rbf} kernel and $nu=0.5$ fitted to all examples provided in the offline phase. This baseline is unable to classify among known classes and distinguish unknown types.

    \item Confidence-based Novelty Classifier (\textsc{cnd}) -- a baseline employing the confidence of the Multilayer Perceptron classifier (\textsc{mlp}) with a single hidden layer of $100$ neurons. The samples will be labeled as unknown if the decision support lies below the threshold of $0.9$. The known classes are recognized with a model if the confidence lies above a threshold. Samples recognized as known are used for self-training. This method is unable to distinguish types of unknowns.

\end{itemize}

Most of the parameters of the reference methods were left as default according to the source library. Some of them in ECSMiner and \textsc{echo} were reduced to fit the size of the offline phase. The hyperparameters of clustering methods were adjusted to the feature scale of each dataset, and for CluStream, a true number of classes was provided.

In experiments, the offline phase length of \textsc{orcca} and all other methods that employ offline training was set depending on the number of considered known classes in synthetic and semi-synthetic data. For each known class, 125 samples were considered, making the offline phase last 250, 500, or 750 samples for tabular and \textsc{cifar} base distributions. In real-world data streams, the lengths of offline phases were 1,500 samples for Covertype and 2,000 for \textsc{kdd} and \textsc{pamap2}. In the cases of \textsc{kdd} and \textsc{pamap2}, only one class was present in the offline phase.

\subsection{Evaluation criteria}

The evaluation metrics were based on Balanced Accuracy (\textsc{bac}) and Adjusted Rand Index (\textsc{ari}). \textsc{Bac} is presented in Equation (\ref{eq:bac}), where $Rec_c$ indicates recall for class $c$ and $N$ denotes the number of classes. \textsc{Bac} will be $1/N$ when random predictions are made and $1$ for perfect recognition.

\begin{equation}
\mathrm{BAC} = \frac{1}{N} \sum_{c = 1, \ldots, N} \mathrm{Rec}_c
\label{eq:bac}
\end{equation}

Balanced accuracy, as a classification metric, was selected due to its symmetry when faced with various levels of class proportions in imbalanced cases \cite{brzezinski2019dynamics}. In dynamic environments, especially when the dichotomy between knowns and unknowns is considered, the proportions of classes can change significantly. \textsc{Bac} will reliably describe scenarios when known samples represent the majority of samples or when novelties significantly outnumber knowns.

Rand Index (\textsc{ri}) is shown in Equation (\ref{eq:ri}), where $N_{11}$ denotes the number of pairs samples with the same label that fell into the same cluster, and $N_{00}$ denotes the number of pairs of samples with different labels that fell into different clusters. The denominator is the number of all possible unordered pairs.

\begin{equation}
\mathrm{RI} = \frac{N_{11} + N_{00}}{\binom{n}{2}}
\label{eq:ri}
\end{equation}

The adjusted version of the metric in Equation (\ref{eq:ari}) uses $\mathrm{RI}_{E}$, which denotes the expected value of \textsc{ri} in random cluster assignment. \textsc{Ari} obtains a value close to $0$ for random assignment, and $1$ for perfect grouping.

\begin{equation}
\mathrm{ARI} = \frac{\mathrm{RI} - \mathrm{RI}_{E}}{1 - \mathrm{RI}_{E}}
\label{eq:ari}
\end{equation}

The three criteria used in experiments to evaluate the quality are defined as follows:
\begin{itemize}
    \item Known class balanced accuracy (\textsc{kc bac}) -- a classification quality measured only on known-class samples in the online phase. A known sample labeled as unknown will decrease the accuracy. In this metric, $N$ is equal to the number of known classes.
    \item Known vs unknown class balanced accuracy  (\textsc{kc/uc bac}) -- a dichotomy between known and unknown samples. Measuring the ability of methods to recognize unknowns. In this metric, $N=2$. All known classes are grouped into the positive class, and all unknowns are grouped into the negative class.
    \item Unknown class Adjusted Rand Index (\textsc{uc ari}) -- a separation of unknowns into individual classes without supervision.
\end{itemize}

The quality was measured for the entire data stream. In cases where a specific measure could not be calculated for a given dataset or model -- e.g. due to only one known class in real-world data streams -- it was skipped and not analyzed. Otherwise, all three criteria were treated as equally important in the evaluation. Since clustering algorithms used as the reference approaches arbitrarily select the cluster labels of known classes (fully unsupervised offline phase), the evaluation used the optimal cluster-to-class matching \cite{han2019learning}. Such matching is performed only for clustering-based algorithms. For methods trained with the labeled set, the predicted labels are directly compared with the ground-truth.

\section{Results of experiments}

This section presents and analyzes the results of experiments. First, the quality of methods will be compared in three criteria. Next, the processing time and memory usage will be studied. The final experiment is an ablation study of \textsc{orcca}.

\subsection{Recognition quality}
The results for synthetic tabular data streams are shown in Tables \ref{tab:tabular_c1} to \ref{tab:tabular_c3}, presenting known class balanced accuracy (criterion~1),  known vs unknown class balanced accuracy (criterion~2), and unknown class \textsc{ari} (criterion~3), respectively. Rows indicate data stream configuration, with K and N denoting the number of known and novel classes, and D the number of concept drifts. The last row of each table shows the mean rank of a method in 15 stream configurations, averaged across replications. Critical Difference (\textsc{cd}) was calculated using the Nemenyi post-hoc test \cite{demvsar2006statistical} at $\alpha = 0.05$. The differences in \textsc{cd} come from the number of methods that could be evaluated in a given criterion.
All tables present results averaged from ten replications. 

\begin{table}[!htb]
    \centering
    \caption{Results of known class balanced accuracy (criterion~1) metric for tabular data streams (CD=2.33).}
    \setlength{\tabcolsep}{3pt}
\renewcommand{\arraystretch}{1.0}
\scriptsize

\begin{tabular}{l|rrrrrrrr}
\toprule
 \textsc{stream}       &   ORCCA &   MINAS &   ECSMin &   ECHO &   CluStr &   DenStr & OCND   &   CND \\
\midrule
 K2 N8 D0 &  0.974 &   \textbf{0.997} &        0.961 &  0.927 &       0.184 &       0.815 & ---    & 0.964 \\
 K2 N8 D1 &  0.975 &   \textbf{0.979} &        0.908 &  0.927 &       0.185 &       0.807 & ---    & 0.966 \\
 K2 N8 D3 &  \textbf{0.986} &   0.679 &        0.639 &  0.936 &       0.206 &       0.817 & ---    & 0.973 \\
 K2 N8 D5 &  \textbf{0.991} &   0.470  &        0.446 &  0.95  &       0.25  &       0.75  & ---    & 0.990  \\
 K2 N8 D7 &  \textbf{0.991} &   0.359 &        0.309 &  0.931 &       0.28  &       0.755 & ---    & 0.980  \\ \midrule
 K4 N6 D0 &  0.966 &   \textbf{0.992} &        0.97  &  0.72  &       0.182 &       0.806 & ---    & 0.908 \\
 K4 N6 D1 &  \textbf{0.968} &   0.957 &        0.912 &  0.72  &       0.192 &       0.805 & ---    & 0.908 \\
 K4 N6 D3 &  \textbf{0.983} &   0.606 &        0.615 &  0.768 &       0.229 &       0.843 & ---    & 0.886 \\
 K4 N6 D5 &  \textbf{0.986} &   0.408 &        0.408 &  0.751 &       0.265 &       0.877 & ---    & 0.947 \\
 K4 N6 D7 &  \textbf{0.982} &   0.292 &        0.277 &  0.806 &       0.33  &       0.833 & ---    & 0.914 \\ \midrule
 K6 N2 D0 &  0.973 &   \textbf{0.991} &        0.786 &  0.548 &       0.181 &       0.793 & ---    & 0.874 \\
 K6 N2 D1 &  \textbf{0.980}  &   0.894 &        0.732 &  0.543 &       0.205 &       0.772 & ---    & 0.847 \\
 K6 N2 D3 &  \textbf{0.991} &   0.423 &        0.431 &  0.518 &       0.297 &       0.807 & ---    & 0.895 \\
 K6 N2 D5 &  \textbf{0.998} &   0.286 &        0.257 &  0.543 &       0.352 &       0.793 & ---    & 0.973 \\
 K6 N2 D7 &  \textbf{1.000}     &   0.318 &        0.249 &  0.522 &       0.36  &       0.556 & ---    & 0.951 \\ \midrule
  \textsc{mean rank} &  \textbf{1.33}  &   3.87  &        5.27  &  4.4   &       6.6   &       4     & ---    & 2.53  \\
\bottomrule
\end{tabular}
    \label{tab:tabular_c1}
\end{table}

Table \ref{tab:tabular_c1} presents balanced accuracy score of recognizing known instances. The tabular data streams present a relatively simple classification problem, hence, several  methods obtain scores above 95\%, especially when fewer than three concept drifts occur. The results also show that when more known classes are introduced, the task becomes more complex for ECSMiner and \textsc{echo}. For static data streams, \textsc{minas} shows the best quality. However, when concept drifts are introduced, its scores drop to less than 30\%. For the data streams with concept drifts, \textsc{orcca} has the highest quality, giving it the best (lowest) mean rank. Interestingly, a simple \textsc{cnd} also achieves good scores and can adapt to drifting distributions. The adaptation abilities result from using the classifier's predictions to incrementally train \textsc{mlp}. With smooth concept drifts, as in this experiment, such a strategy gives \textsc{cnd} a mean rank within a critical distance from the leading \textsc{orcca}. 

\begin{table}[!htb]
    \centering
    \caption{Results of known vs unknown class balanced accuracy (criterion~2) metric for tabular data streams (CD=2.71).}
    \setlength{\tabcolsep}{3pt}
\renewcommand{\arraystretch}{1.0}
\scriptsize

\begin{tabular}{l|rrrrrrrr}
\toprule
 \textsc{stream}       &   ORCCA &   MINAS &   ECSMin &   ECHO &   CluStr &   DenStr & OCND   &   CND \\
\midrule
 K2 N8 D0 &  \textbf{0.933} &   0.831 &        0.706 &  0.505 &       0.498 &       0.769 &  0.705 & 0.509 \\
 K2 N8 D1 &  \textbf{0.947} &   0.833 &        0.680  &  0.501 &       0.504 &       0.767 &  0.674 & 0.511 \\
 K2 N8 D3 &  \textbf{0.982} &   0.689 &        0.544 &  0.496 &       0.499 &       0.764 &  0.579 & 0.511 \\
 K2 N8 D5 &  \textbf{0.989} &   0.590  &        0.448 &  0.490  &       0.499 &       0.783 &  0.536 & 0.510  \\
 K2 N8 D7 &  \textbf{0.991} &   0.536 &        0.378 &  0.480  &       0.506 &       0.798 &  0.511 & 0.505 \\ \midrule
 K4 N6 D0 &  \textbf{0.893} &   0.794 &        0.581 &  0.501 &       0.500   &       0.536 &  0.641 & 0.508 \\
 K4 N6 D1 &  \textbf{0.912} &   0.791 &        0.551 &  0.498 &       0.500   &       0.534 &  0.596 & 0.511 \\
 K4 N6 D3 &  \textbf{0.969} &   0.622 &        0.402 &  0.490  &       0.500   &       0.512 &  0.492 & 0.514 \\
 K4 N6 D5 &  \textbf{0.980}  &   0.530  &        0.296 &  0.480  &       0.500   &       0.556 &  0.449 & 0.516 \\
 K4 N6 D7 &  \textbf{0.982} &   0.465 &        0.230  &  0.470  &       0.500   &       0.521 &  0.427 & 0.519 \\ \midrule
 K6 N2 D0 &  \textbf{0.846} &   0.714 &        0.518 &  0.500   &       0.500   &       0.561 &  0.597 & 0.518 \\
 K6 N2 D1 &  \textbf{0.904} &   0.705 &        0.488 &  0.497 &       0.500   &       0.556 &  0.511 & 0.514 \\
 K6 N2 D3 &  \textbf{0.967} &   0.513 &        0.303 &  0.491 &       0.500   &       0.581 &  0.392 & 0.513 \\
 K6 N2 D5 &  \textbf{0.976} &   0.412 &        0.183 &  0.489 &       0.500   &       0.523 &  0.352 & 0.517 \\
 K6 N2 D7 &  \textbf{0.950}  &   0.399 &        0.159 &  0.482 &       0.500   &       0.697 &  0.308 & 0.515 \\
\midrule
 \textsc{mean rank} &  \textbf{1}     &   3.13  &        6.6   &  6.73  &       5.87  &       2.87  &  5.13  & 4.67  \\
\bottomrule
\end{tabular}
    \label{tab:tabular_c2}
\end{table}

Table \ref{tab:tabular_c2} shows the results of binary classification between knowns and unknowns, i.e., the detection of novelties. The random classification \textsc{bac} in this setting is 50\%. Due to concept drifts, the scores of many methods drop even below this value. \textsc{Orcca} is the only method that can consistently detect unknowns and, therefore, achieves the best mean rank. \textsc{Minas} works well when the concepts are static or in cases with a single concept drift, but its quality again drops with more frequent concept changes. DenStream achieves high scores when streams have 2 known classes and 8 unknown ones (top section), but when more known classes are present, its quality only slightly exceeds 50\%.

\begin{table}[!htb]
    \centering
    \caption{Results of unknown class ARI (criterion~3) metric for tabular streams (CD=1.95).}
    \setlength{\tabcolsep}{3pt}
\renewcommand{\arraystretch}{1.0}
\scriptsize

\begin{tabular}{l|rrrrrrrr}
\toprule
 \textsc{stream}       &   ORCCA &   MINAS &   ECSMin &   ECHO &   CluStr &   DenStr & OCND   &   CND \\
\midrule
 K2 N8 D0 &  0.443 &   0.276 &        0.051 &  0.033 &       0.026 &       \textbf{0.611} & ---    & ---   \\
 K2 N8 D1 &  0.458 &   0.295 &        0.049 &  0.039 &       0.026 &       \textbf{0.617} & ---    & ---   \\
 K2 N8 D3 &  0.489 &   0.304 &        0.047 &  0.038 &       0.028 &       \textbf{0.526} & ---    & ---   \\
 K2 N8 D5 &  \textbf{0.483} &   0.313 &        0.039 &  0.033 &       0.028 &       0.479 & ---    & ---   \\
 K2 N8 D7 &  \textbf{0.495} &   0.345 &        0.041 & -0.002 &       0.028 &       0.442 & ---    & ---   \\ \midrule
 K4 N6 D0 &  0.378 &   0.317 &        0.070  &  0.19  &       0.024 &       \textbf{0.579} & ---    & ---   \\
 K4 N6 D1 &  0.398 &   0.324 &        0.068 &  0.218 &       0.024 &       \textbf{0.586} & ---    & ---   \\
 K4 N6 D3 &  0.461 &   0.324 &        0.069 &  0.086 &       0.027 &       \textbf{0.464} & ---    & ---   \\
 K4 N6 D5 &  \textbf{0.480} &   0.358 &        0.067 & -0.004 &       0.027 &       0.358 & ---    & ---   \\
 K4 N6 D7 &  \textbf{0.474} &   0.357 &        0.068 & -0.003 &       0.031 &       0.395 & ---    & ---   \\ \midrule
 K6 N2 D0 &  0.210  &   0.216 &        0.046 &  0.348 &       0.009 &       \textbf{0.745} & ---    & ---   \\
 K6 N2 D1 &  0.215 &   0.187 &        0.046 & -0.018 &       0.006 &       \textbf{0.668} & ---    & ---   \\
 K6 N2 D3 &  \textbf{0.617} &   0.335 &        0.046 & -0.035 &       0.011 &       0.374 & ---    & ---   \\
 K6 N2 D5 &  \textbf{0.697} &   0.231 &        0.058 & -0.027 &       0.018 &       0.119 & ---    & ---   \\
 K6 N2 D7 &  \textbf{0.414} &   0.225 &        0.056 & -0.018 &       0.025 &       0.142 & ---    & ---   \\ \midrule
\textsc{mean rank} &  1.67  &   2.87  &        4.27  &  5.07  &       5.53  &       \textbf{1.6}   & ---    & ---   \\
\bottomrule
\end{tabular}
    \label{tab:tabular_c3}
\end{table}

Table \ref{tab:tabular_c3} presents \textsc{ari} of unknown class grouping. This fully unsupervised task is more complex than the previous ones. When fewer than three concept drifts are present, DenStream achieves the highest \textsc{ari} of 58\% to 75\%. For streams with many drifts (D5 or D7), the highest \textsc{ari} is observed for \textsc{orcca} (41\% to 70\%). \textsc{Minas} method usually achieves results between 20 and 30\%. The remaining methods are unable to correctly distinguish unknowns. The best mean ranks are achieved by DenStream and \textsc{orcca}. 

\begin{table}[!htb]
    \centering
    \caption{Results of known class balanced accuracy (criterion~1) metric for CIFAR data streams (CD=2.33).}
    \setlength{\tabcolsep}{3pt}
\renewcommand{\arraystretch}{1.0}
\scriptsize

\begin{tabular}{l|rrrrrrrr}
\toprule
 \textsc{stream}       &   ORCCA &   MINAS &   ECSMin &   ECHO &   CluStr &   DenStr & OCND   &   CND \\
\midrule
 K2 N8 D0 &  0.732 &   \textbf{0.981} &        0.962 &  0.952 &       0.424 &       0.434 & ---    & 0.799 \\
 K2 N8 D1 &  0.720  &   \textbf{0.980}  &        0.932 &  0.955 &       0.421 &       0.436 & ---    & 0.805 \\
 K2 N8 D3 &  0.819 &   0.893 &        0.735 &  \textbf{0.947} &       0.432 &       0.466 & ---    & 0.864 \\
 K2 N8 D5 &  0.924 &   0.732 &        0.553 &  \textbf{0.937} &       0.429 &       0.559 & ---    & 0.889 \\
 K2 N8 D7 &  \textbf{0.964} &   0.569 &        0.410  &  0.926 &       0.408 &       0.695 & ---    & 0.922 \\
 \midrule
 K4 N6 D0 &  0.548 &   \textbf{0.881} &        0.796 &  0.630  &       0.258 &       0.408 & ---    & 0.387 \\
 K4 N6 D1 &  0.531 &   \textbf{0.880}  &        0.767 &  0.628 &       0.258 &       0.411 & ---    & 0.387 \\
 K4 N6 D3 &  0.692 &   \textbf{0.768} &        0.563 &  0.629 &       0.263 &       0.535 & ---    & 0.430  \\
 K4 N6 D5 &  \textbf{0.781} &   0.560  &        0.389 &  0.682 &       0.256 &       0.604 & ---    & 0.474 \\
 K4 N6 D7 &  \textbf{0.864} &   0.419 &        0.270  &  0.694 &       0.254 &       0.614 & ---    & 0.502 \\
 \midrule
 K6 N2 D0 &  0.528 &   \textbf{0.850}  &        0.634 &  0.490  &       0.174 &       0.369 & ---    & 0.249 \\
 K6 N2 D1 &  0.576 &   \textbf{0.844} &        0.612 &  0.496 &       0.175 &       0.376 & ---    & 0.251 \\
 K6 N2 D3 &  \textbf{0.754} &   0.604 &        0.388 &  0.519 &       0.174 &       0.504 & ---    & 0.357 \\
 K6 N2 D5 &  0.957 &   0.412 &        0.211 &  0.517 &       0.176 &       0.465 & ---    & \textbf{0.971} \\
 K6 N2 D7 &  0.990  &   0.417 &        0.209 &  0.537 &       0.175 &       0.223 & ---    & \textbf{0.991} \\
 \midrule
\textsc{mean rank}    &  2.67  &   \textbf{2.53}  &        4.2   &  2.6   &       7     &       4.73  & ---    & 4.27  \\

\bottomrule
\end{tabular}
    \label{tab:cifar_c1}
\end{table}

The results for \textsc{cifar}, presented in Tables \ref{tab:cifar_c1} to \ref{tab:cifar_c3}, show similar trends, with the original sample distribution being more challenging. 

The significant benefit of the \textsc{orcca} adaptation mechanism compared to references becomes clearly visible in criterion~1, shown in Table \ref{tab:cifar_c1}. When data streams have more concept drifts, the proposed approach ranks better. Interestingly, the simple procedure employed in \textsc{cnd} performs best when five or more drifts are involved in N2 streams. This is likely linked to a large number of known classes (K6) classified with high confidence, that enables the method to adapt to concept drifts. It is worth mentioning that a smooth concept drift mechanism implemented in the \textsc{owdsg} generator translates the class distribution in a random direction, often moving classes away from the initial decision boundary (\textit{virtual} drift) and increasing the separation of distributions when the method is able to adapt to such shift i.e., follow the class.

\begin{table}[!htb]
    \centering
    \caption{Results of known vs unknown class balanced accuracy (criterion~2) for CIFAR data streams (CD=2.71).}
    \setlength{\tabcolsep}{3pt}
\renewcommand{\arraystretch}{1.0}
\scriptsize

\begin{tabular}{l|rrrrrrrr}
\toprule
 \textsc{stream}       &   ORCCA &   MINAS &   ECSMin &   ECHO &   CluStr &   DenStr & OCND   &   CND \\
\midrule
 K2 N8 D0 &  0.575 &   0.551 &        0.579 &  0.503 &       0.549 &       0.630  &  \textbf{0.653} & 0.488 \\
 K2 N8 D1 &  0.575 &   0.550  &        0.564 &  0.499 &       0.563 &       \textbf{0.631} &  0.620  & 0.489 \\
 K2 N8 D3 &  \textbf{0.737} &   0.511 &        0.462 &  0.492 &       0.556 &       0.607 &  0.521 & 0.492 \\
 K2 N8 D5 &  \textbf{0.921} &   0.429 &        0.367 &  0.484 &       0.555 &       0.563 &  0.479 & 0.493 \\
 K2 N8 D7 &  \textbf{0.971} &   0.345 &        0.295 &  0.478 &       0.593 &       0.566 &  0.452 & 0.500   \\ \midrule

 K4 N6 D0 &  0.517 &   0.518 &        0.529 &  0.501 &       0.484 &       0.532 &  \textbf{0.597} & 0.472 \\
 K4 N6 D1 &  0.508 &   0.520  &        0.509 &  0.498 &       0.514 &       0.532 &  \textbf{0.550}  & 0.472 \\
 K4 N6 D3 &  \textbf{0.676} &   0.459 &        0.379 &  0.491 &       0.513 &       0.508 &  0.443 & 0.473 \\
 K4 N6 D5 &  \textbf{0.782} &   0.342 &        0.270  &  0.480  &       0.488 &       0.493 &  0.402 & 0.480  \\
 K4 N6 D7 &  \textbf{0.850}  &   0.259 &        0.198 &  0.471 &       0.496 &       0.499 &  0.381 & 0.484 \\ \midrule

 K6 N2 D0 &  0.504 &   0.504 &        0.508 &  0.499 &       0.542 &       0.543 &  \textbf{0.556} & 0.479 \\
 K6 N2 D1 &  \textbf{0.556} &   0.502 &        0.489 &  0.497 &       0.552 &       0.535 &  0.464 & 0.479 \\
 K6 N2 D3 &  \textbf{0.678} &   0.370  &        0.311 &  0.492 &       0.547 &       0.505 &  0.350  & 0.480  \\
 K6 N2 D5 &  \textbf{0.929} &   0.243 &        0.180  &  0.489 &       0.525 &       0.502 &  0.311 & 0.500   \\
 K6 N2 D7 &  \textbf{0.993} &   0.218 &        0.156 &  0.482 &       0.498 &       0.475 &  0.256 & 0.506 \\ \midrule
\textsc{mean rank} &  \textbf{2.2}  &   5.8   &        6.47  &  5.53  &       3.33  &       2.47  &  4.53  & 5.67  \\

\bottomrule
\end{tabular}
    \label{tab:cifar_c2}
\end{table}

Novelty detection in \textsc{cifar} data is a more challenging task. The results in Table \ref{tab:cifar_c2} show that in stationary settings (no concept drifts) classic novelty detectors (\textsc{minas}, ECSMiner, \textsc{echo}) do not perform well. The best recognition quality in this setting is observed for the DenStream clustering and one-class-based method (\textsc{ocnd}). When concept drifts are involved, \textsc{orcca} achieves the highest quality.

\begin{table}[!htb]
    \centering
    \caption{Results of unknown class ARI (criterion~3) metric for CIFAR streams (CD=1.95).}
    \setlength{\tabcolsep}{3pt}
\renewcommand{\arraystretch}{1.0}
\scriptsize

\begin{tabular}{l|rrrrrrrr}
\toprule
 \textsc{stream}       &   ORCCA &   MINAS &   ECSMin &   ECHO &   CluStr &   DenStr & OCND   &   CND \\
\midrule
 K2 N8 D0 &  0.167 &   0.153 &        0.152 &  0.142 &       0.059 &       \textbf{0.306} & ---    & ---   \\
 K2 N8 D1 &  0.159 &   0.153 &        0.152 &  0.145 &       0.059 &       \textbf{0.305} & ---    & ---   \\
 K2 N8 D3 &  0.084 &   0.156 &        0.149 &  0.132 &       0.056 &       \textbf{0.316} & ---    & ---   \\
 K2 N8 D5 &  0.075 &   0.156 &        0.150  &  0.069 &       0.058 &       \textbf{0.315} & ---    & ---   \\
 K2 N8 D7 &  0.101 &   0.155 &        0.152 &  0.058 &       0.068 &       \textbf{0.319} & ---    & ---   \\ \midrule

 K4 N6 D0 &  0.249 &   0.255 &        0.238 &  0.215 &       0.067 &       \textbf{0.368} & ---    & ---   \\
 K4 N6 D1 &  0.241 &   0.257 &        0.238 &  0.219 &       0.063 &       \textbf{0.369} & ---    & ---   \\
 K4 N6 D3 &  0.113 &   0.254 &        0.237 &  0.137 &       0.062 &       \textbf{0.383} & ---    & ---   \\
 K4 N6 D5 &  0.121 &   0.257 &        0.237 &  0.047 &       0.053 &       \textbf{0.410}  & ---    & ---   \\
 K4 N6 D7 &  0.151 &   0.256 &        0.238 &  0.030  &       0.057 &       \textbf{0.386} & ---    & ---   \\ \midrule

 K6 N2 D0 &  0.232 &   0.223 &        0.226 &  \textbf{0.309} &       0.077 &       0.138 & ---    & ---   \\
 K6 N2 D1 &  \textbf{0.242} &   0.223 &        0.226 &  0.193 &       0.04  &       0.139 & ---    & ---   \\
 K6 N2 D3 &  \textbf{0.271} &   0.221 &        0.226 &  0.139 &       0.095 &       0.129 & ---    & ---   \\
 K6 N2 D5 &  \textbf{0.448} &   0.241 &        0.228 & -0.024 &       0.077 &       0.135 & ---    & ---   \\
 K6 N2 D7 &  \textbf{0.576} &   0.222 &        0.268 & -0.025 &       0.092 &       0.077 & ---    & ---   \\ \midrule 
  \textsc{mean rank} &  2.8   &   2.47  &        3.07  &  4.8   &       5.6   &       \textbf{2.27}  & ---    & ---   \\

\bottomrule
\end{tabular}
    \label{tab:cifar_c3}
\end{table}

The final table for \textsc{cifar} (Table \ref{tab:cifar_c3}) shows \textsc{uc ari} in \textsc{cifar} data streams. In this setting, DenStream clustering algorithm is a clear leader, especially when 6 and 8 novelty groups are observed. When only two novel classes appear (bottom section of a table), the task becomes easier for all novelty detectors, with \textsc{orcca} leading in 4 out of 5 stream types. Even though \textsc{minas} approach does not achieve supreme quality, its mean rank is low, showing its consistent ability to distinguish unknowns. \textsc{Orcca} ranks third in this evaluation criterion. Its quality increases when the drifts are introduced in streams with 6 known classes. The loss of quality when fewer known classes are observed, even though criterion~2 showed the ability to distinguish knowns and unknowns, may indicate that the method confuses groups of unknowns when their number is significant.

\begin{figure}[!htb]
    \centering
    \includegraphics[width=\linewidth]{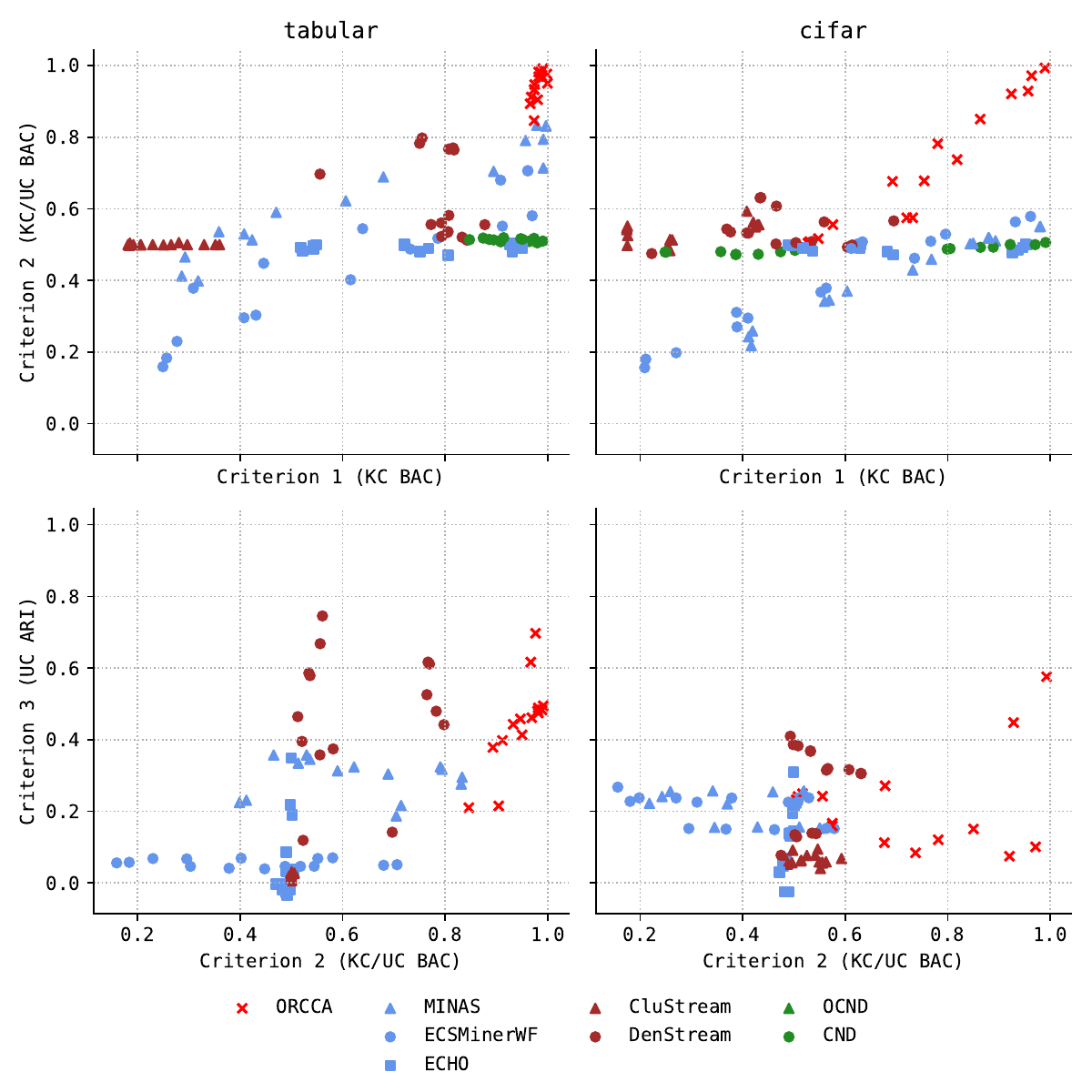}
    \caption{Averaged results from comparison experiment on synthetic and semi-synthetic data streams (tabular and based on CIFAR) in all three criteria, grouped in pairs.}
    \label{fig:compare}
\end{figure}

Metric values for both source distributions used in constructing streams are shown in Figure \ref{fig:compare}. The figure shows pairs of metrics in the three-criterion evaluation -- criteria~1 and~2 in the top row, and criteria~2 and 3 in the bottom row. The mean results of \textsc{orcca} are indicated with red markers, which are grouped towards the top-right corner of each plot, indicating good performance. The results of \textsc{cifar}-based streams explicitly show that only \textsc{orcca} has a \textsc{kc/uc bac} (criterion~2) that consistently exceeds 60\%.

\begin{figure}[!htb]
    \centering
    \includegraphics[width=\linewidth]{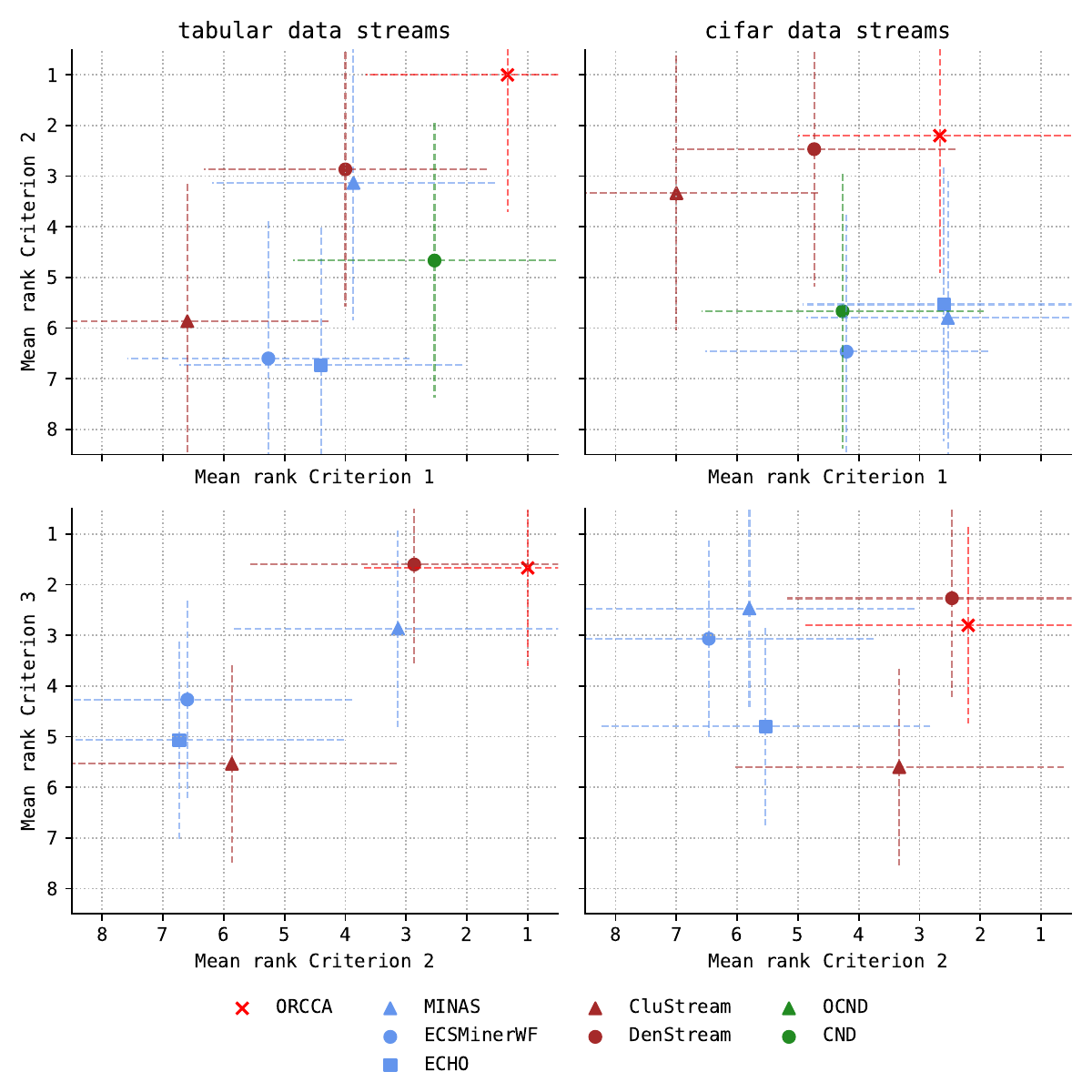}
    \caption{The presentation of ranks of all evaluated methods and the critical differences (dashed lines) in all three criteria, grouped in pairs. The best methods will appear in top-right corner of plots.}
    \label{fig:compare-rank}
\end{figure}

Additionally, ranks are presented in Figure \ref{fig:compare-rank} in a similar manner. The ranges of \textsc{cd} are indicated with dashed lines. 

The same experiment was conducted for real-world data streams. The results are shown in Table \ref{tab:real} for all three criteria. 

\begin{table}[!htb]
    \centering
    \caption{Results of quality metrics for real-world data streams in all three criteria. For KDD and PAMAP2 only one class was visible in the offline phase, making it impossible to measure performance in criterion~1.}
    \setlength{\tabcolsep}{3pt}
\renewcommand{\arraystretch}{1.0}
\scriptsize

\begin{tabular}{l|rrrrrrrr}
\toprule
 \textsc{stream}       &   ORCCA &   MINAS &   ECSMin &   ECHO &   CluStr &   DenStr & OCND   &   CND \\
\midrule
     & \multicolumn{8}{c}{\textsc{known class balanced accuracy (criterion 1)}}  \\
\textsc{covertype} & 0.202  & 0.141   & \textbf{0.394}        & 0.325  & 0.207       & 0.341       & ---    & 0.330  \\
 \textsc{kdd}       & ---    & ---     & ---          & ---    & ---         & ---         & ---    & ---   \\
 \textsc{pamap2}    & ---    & ---     & ---          & ---    & ---         & ---         & ---    & ---   \\
\midrule
 \textsc{mean rank} & 6    & 7     & \textbf{1}          & 4    & 5         & 2         & ---    & 3   \\
\midrule
     & \multicolumn{8}{c}{\textsc{known vs unknown class balanced accuracy (criterion 2)}}  \\
\textsc{covertype}          &  \textbf{0.723} &   0.648 &        0.503 &  0.531 &       0.500   &       0.649 &  0.613 & 0.500   \\
 \textsc{kdd}                 &  \textbf{0.857} &   0.647 &        0.504 &  0.665 &       0.411 &       0.408 &  0.641 & ---   \\
 \textsc{pamap2}              &  0.638 &   0.520  &        \textbf{0.651} &  0.507 &       0.568 &       0.402 &  0.500   & ---   \\ 
 \midrule
 \textsc{mean rank}  &  \textbf{1.33}  &   3.33  &        4     &  4     &       5.67  &       5.33  &  4.67  & 7   \\
\midrule
     & \multicolumn{8}{c}{\textsc{unknown class ARI (criterion 3)}}  \\
\textsc{covertype}           &  \textbf{0.162} &   0.070  &        0.001 & -0.002 &       0.012 &       0.057 & ---    & ---   \\
 \textsc{kdd}               &  \textbf{0.935} &   0.260  &        0.040  &  0.400   &       0.692 &      -0.005 & ---    & ---   \\
 \textsc{pamap2}              &  \textbf{0.095} &   0.032 &        0.026 &  0.049 &       0.068 &       0.015 & ---    & ---   \\
 \midrule
 \textsc{mean rank} &  \textbf{1}     &   3.33  &        5     &  4     &       2.67  &       5     & ---    & ---   \\
\bottomrule
\end{tabular}
    \label{tab:real}
\end{table}

More than one known class was present only in the Covertype data stream, therefore, known class \textsc{bac} was measured for this dataset only. The highest score in this criterion was achieved by ECSMiner, however, only two methods have slightly exceeded a random-assignment score in three-class classification (33\%). \textsc{Orcca} did not perform well in this criterion, reaching only 0.202 \textsc{bac}. Such a poor ability to recognize among knowns may be linked to classifying known samples as novelties, likely due to a sudden concept drift. The complexity of recognition between known classes in this dataset may also be linked to class imbalance. The proportions of classes in the offline phase differ significantly from their proportions online, since the order of samples results from their spatial location. If the known class is rare in the online phase, its recognition will still weigh $1/N$ of the final balanced accuracy.

For novelty detection (criterion~2) and grouping of unknowns (criterion~3), \textsc{orcca} obtained the highest scores, with a single exception where ECSMiner marginally surpassed proposed method in criterion 2 on the \textsc{pamap2} data stream. In the remaining data streams, the advantage of the proposed method was substantial. 

\subsection{Time and Memory Complexity}

The time and memory complexity experiment was conducted using a 14-core \textsc{cpu} with an \textsc{arm} architecture and a maximum clock rate of 4.51 GHz. The time complexity of \textsc{orcca} is bounded by $O(M(dh+hb))$, with $M$ being the sum of known classes and recognized novelties. The equation assumes a single hidden layer in the encoder and decoder, as in the default implementation. The model size will therefore depend on the dimensionality of the data $d$, the number of hidden neurons $h$, and the bottleneck $b$. This bound treats the buffer size $B$ and the number of training epochs as constants.

The results are presented in Tables \ref{tab:time} and \ref{tab:memory}. The tables show 6 types of streams and the values measured on the initial and final 25,000 samples of the online phase, averaged over 5 stream replications. Due to significant computation time, metrics for CluStream were measured for only one replication. Measuring the metrics at the beginning and the end of processing allowed for estimating the increase in processing time and the growth of the model state, which is included in the tables. Negative values in this row indicate that the method processed samples faster or used less memory in the end of a stream.

\begin{table}[!htb]
    \centering
    \caption{Processing time per sample (in $\mu s$) by the proposed and the reference methods.}
    \setlength{\tabcolsep}{3pt}
\renewcommand{\arraystretch}{1.0}
\scriptsize

\begin{tabular}{l|rrrrrrrr}
\toprule
 \textsc{blocks}       &   ORCCA &   MINAS &   ECSMin &   ECHO &   CluStr &   DenStr & OCND   &   CND \\
\midrule
     & \multicolumn{8}{c}{\textsc{stream: 0 novelties, 0 drifts}}  \\
     
\textsc{start}    &  435.2 &     7.7 &         43.3 & 1537.0   &      3931.1 &       102.2 &   42.4 & 498.6 \\
\textsc{end}     &  434.4 &     7.8 &         43.1 & 1558.9 &      3882.2 &       102.0   &   42.1 & 506.4 \\
\textsc{\% growth} &   -0.2 &     0.4 &         -0.5 &    1.4 &        -1.2 &        -0.1 &   -0.5 &   1.6 \\

\midrule
& \multicolumn{8}{c}{\textsc{stream: 0 novelties, 4 drifts}}  \\
\textsc{start}    &  437.0   &     8.6 &         66.7 & 1431.4 &      3797.6 &       102.1 &   42.1 & 531.9 \\
\textsc{end}      &  440.9 &     8.0   &        245.6 & 1397.9 &      4056.5 &        70.8 &   42.4 & 525.5 \\
\textsc{\% growth} &    0.9 &    -7.2 &        268.2 &   -2.3 &         6.8 &       -30.7 &    0.8 &  -1.2 \\

\midrule
& \multicolumn{8}{c}{\textsc{stream: 3 novelties, 0 drifts}}  \\
\textsc{start}    &  436.1 &     8.0   &         50.0   & 1944.0   &      3957.0   &       131.7 &   41.9 & 524.4 \\
\textsc{end}     &  451.8 &     8.4 &         78.5 & 2035.1 &      4024.9 &       222.6 &   41.9 & 530.0   \\
\textsc{\% growth} &    3.6 &     5.3 &         57.0   &    4.7 &         1.7 &        69.0   &    0.1 &   1.1 \\

\midrule
& \multicolumn{8}{c}{\textsc{stream: 3 novelties, 4 drifts}}  \\
\textsc{start}    &  433.1 &     8.7 &         70.5 & 2056.2 &      3737.5 &       129.2 &   41.6 & 522.3 \\
\textsc{end}      &  462.1 &    10.3 &        215.0   & 1353.1 &      4101.2 &       187.0   &   41.6 & 528.9 \\
\textsc{\% growth} &    6.7 &    18.5 &        205.1 &  -34.2 &         9.7 &        44.8 &    0.0   &   1.3 \\

\midrule
& \multicolumn{8}{c}{\textsc{stream: 5 novelties, 0 drifts}}  \\
\textsc{start}    &  432.0   &     8.2 &         75.2 & 1508.3 &      4000.1 &       136.7 &   41.8 & 521.3 \\
\textsc{end}      &  455.3 &     9.2 &        152.7 & 2097.5 &      4119.9 &       198.7 &   41.8 & 528.0   \\
\textsc{\% growth} &    5.4 &    12.3 &        102.9 &   39.1 &         3.0   &        45.3 &    0.1 &   1.3 \\

\midrule
& \multicolumn{8}{c}{\textsc{stream: 5 novelties, 4 drifts}}  \\
\textsc{start}    &  432.2 &     9.0   &        100.1 & 1456.2 &      3815.0   &       138.9 &   41.9 & 523.7 \\
\textsc{end}     &  508.2 &    10.9 &        272.1 & 1570.7 &      4053.2 &       339.5 &   41.7 & 528.0   \\
\textsc{\% growth} &   17.6 &    20.9 &        171.7 &    7.9 &         6.2 &       144.5 &   -0.4 &   0.8 \\
\hline
\end{tabular}
    \label{tab:time}
\end{table}

Table \ref{tab:time} shows that \textsc{minas}, ECSMiner, and \textsc{ocnd} have the lowest processing times. It is worth noting that ECSMiner's result is substantially higher even at the start when the number of novelties and drifts is large. This arises from the fact that in the most non-stationary stream (5 novelties, 4 drifts), both novelty and concept drift appear in the initial 25,000 samples, which may impact the components of this method. \textsc{Echo} and CluStream show the largest processing times. \textsc{Orcca} and \textsc{cnd} times are consistently in the middle range, with 430 to 530 $\mu s$ per sample. In a streaming setting, however, the growth of processing time may be more important than the time itself \cite{domingos2003general}. The streams with more novelties and concept drifts are linked to a greater increase in processing time. The most significant increase is observed for ECSMiner (up to 268\%) and DenStream (up to 144\%). \textsc{Orcca}'s largest observed growth was 17\%.

\begin{table}[!htb]
    \centering
    \caption{Results of memory usage (in $kB$) by the proposed and the reference methods.}
    \setlength{\tabcolsep}{3pt}
\renewcommand{\arraystretch}{1.0}
\scriptsize

\begin{tabular}{l|rrrrrrrr}
\toprule
 \textsc{blocks}       &   ORCCA &   MINAS &   ECSMin &   ECHO &   CluStr &   DenStr & OCND   &   CND \\
\midrule
     & \multicolumn{8}{c}{\textsc{stream: 0 novelties, 0 drifts}}  \\
     
\textsc{start}    &  663.8 &    19.1 &         30.3 &   43.1 &       996.3 &        69.6 &  159.3 & 329.3 \\
\textsc{end}      &  690.1 &    19.0   &         30.1 &   43.2 &       999.6 &        69.6 &  159.3 & 329.3 \\
\textsc{\% growth} &    4.0   &    -0.6 &         -0.6 &    0.4 &         0.3 &         0.0   &    0.0   &   0.0   \\

  \midrule
     & \multicolumn{8}{c}{\textsc{stream: 0 novelties, 4 drifts}}  \\
     
\textsc{start}   &  657.7 &    21.6 &         33.5 &   43.9 &      1005.4 &        69.6 &  159.3 & 329.3 \\
\textsc{end}      &  664.8 &    35.6 &        103.9 &   42.5 &       995.4 &        52.0   &  159.3 & 329.3 \\
\textsc{\% growth} &    1.1 &    64.3 &        210.4 &   -3.3 &        -1.0   &       -25.2 &    0.0   &   0.0   \\

  \midrule
     & \multicolumn{8}{c}{\textsc{stream: 3 novelties, 0 drifts}}  \\
     
\textsc{start}    &  667.0   &    20.5 &         31.1 &   42.4 &      1009.4 &        84.8 &  158.9 & 329.3 \\
\textsc{end}      &  867.7 &    25.3 &         34.8 &   43.0   &      1003.3 &       111.6 &  158.9 & 329.3 \\
\textsc{\% growth} &   30.1 &    23.4 &         12.0   &    1.5 &        -0.6 &        31.6 &    0.0   &   0.0   \\

  \midrule
     & \multicolumn{8}{c}{\textsc{stream: 3 novelties, 4 drifts}}  \\
     
\textsc{start}    &  661.4 &    21.9 &         33.2 &   42.1 &      1017.8 &        84.3 &  158.9 & 329.3 \\
\textsc{end}      &  894.6 &    40.5 &         71.9 &   43.6 &      1006.7 &       107.2 &  158.9 & 329.3 \\
\textsc{\% growth} &   35.3 &    84.8 &        116.8 &    3.5 &        -1.1 &        27.2 &    0.0   &   0.0   \\

  \midrule
     & \multicolumn{8}{c}{\textsc{stream: 5 novelties, 0 drifts}}  \\
     
\textsc{start}    &  665.0   &    21.4 &         34.0   &   44.8 &      1011.3 &        84.8 &  159.3 & 329.3 \\
\textsc{end}      &  871.9 &    29.0   &         40.6 &   42.7 &      1003.9 &       110.2 &  159.3 & 329.3 \\
\textsc{\% growth} &   31.1 &    35.0   &         19.4 &   -4.7 &        -0.7 &        29.9 &    0.0   &   0.0   \\

  \midrule
     & \multicolumn{8}{c}{\textsc{stream: 5 novelties, 4 drifts}}  \\
     
\textsc{start}    &  662.8 &    23.7 &         37.2 &   44.1 &      1022.0   &        85.6 &  159.3 & 329.3 \\
\textsc{end}      &  949.2 &    43.5 &         65.6 &   43.8 &      1009.7 &       137.9 &  159.3 & 329.3 \\
\textsc{\% growth} &   43.2 &    83.8 &         76.4 &   -0.5 &        -1.2 &        61.1 &    0.0   &   0.0   \\
\hline
\end{tabular}
    \label{tab:memory}
\end{table}

Table \ref{tab:memory} presents the memory usage in the same manner. \textsc{Orcca} shows the second highest memory usage, which results from using per-class autoencoders, the number of which increases when novelty is observed or when the number of known classes is large. The number of autoencoder's parameters also depends on the dimensionality of the data. The highest memory usage in this experiment was observed for CluStream. Even though the raw memory usage of \textsc{orcca} was significant, its memory growth was not as severe as for reference approaches. In the most extreme setting of 5 novelties and 4 concept drifts, \textsc{minas}, ECSMiner, and DenStream showed a growth of 61-83\%, while \textsc{orcca} showed a growth of 43\%. It is also worth noting that the resource consumption of the proposed method grows only with the number of discovered classes, and remains constant when concept drifts occur, without novelties. It is consistent with the theoretically estimated $O(M(dh+hb))$ and is linked to the number of models $M$, remaining constant. Any increase of processing time when concept drifts appear may result from redundant, incorrect recognition of novelty in practical setting.

\begin{figure}[!htb]
    \centering
    \includegraphics[width=\linewidth]{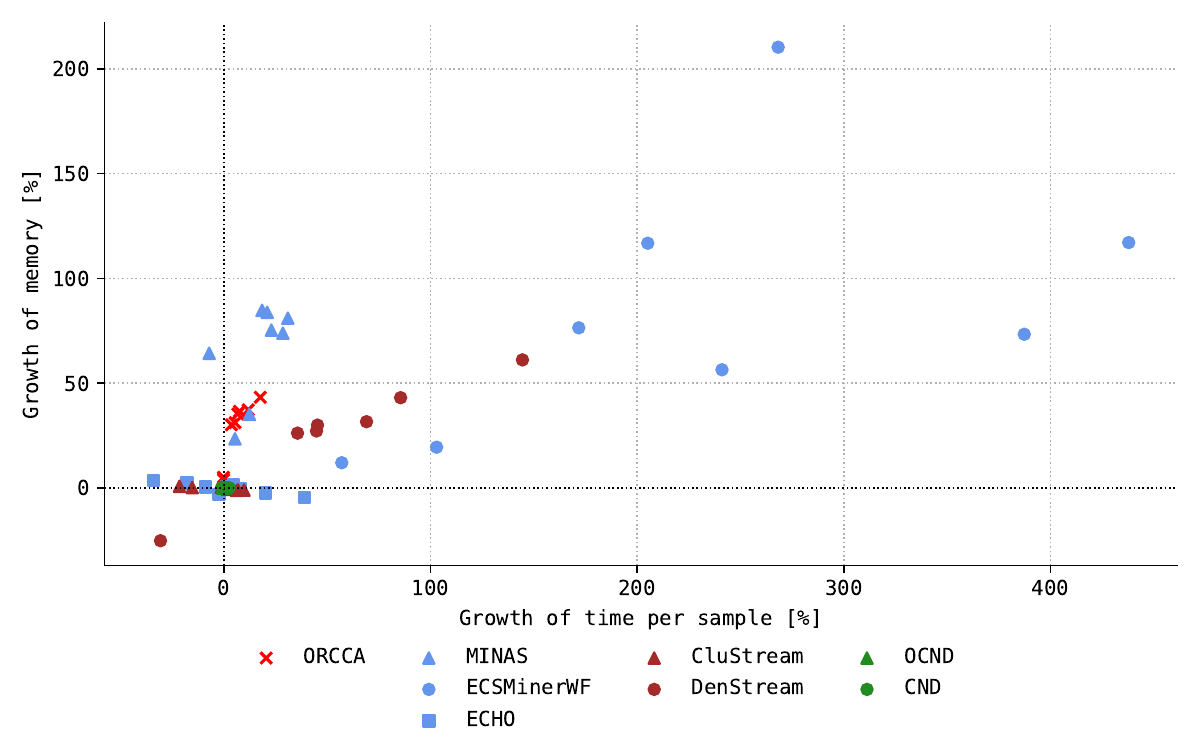}
    \caption{The visual presentation of time and memory resources growth. The points indicate the increase between initial processing steps and at the end of a stream. Red markers indicate results for ORCCA. Its processing time growth is minimal, and its memory growth is lower than that of most reference methods.}
    \label{fig:time-memo}
\end{figure}

To aid the comparison of the methods, the results are also presented in Figure \ref{fig:time-memo}. The points indicate the growth percentage in processing time and memory consumption. The results show, as expected from the tables, that growth in memory usage is a bigger issue compared to processing time. Still, compared to \textsc{minas}, ECSMiner, and DenStream, the resource consumption growth of \textsc{orcca} supports its applicability in a data stream processing setting.

\subsection{Ablation study}

The primary ablation study used the same data streams as in the comparison experiment, limited to tabular distributions. An important part of the method is error bimodality estimation and buffer splitting, meanwhile, this mechanism may not be as essential for synthetic data streams, where one novelty appears at a time. For that reason, the ablation experiment was extended to the Covertype dataset. In total, 150 synthetic data streams were used along with a single real-world one to validate the observations. The results of selected types of streams are presented in Table \ref{tab:ablation}, in all three criteria. 

\begin{table}[!htb]
    \centering
    \caption{Ablation study results. Positive values indicate a benefit of a given mechanism, the negative values indicate a reduction of recognition quality in relation to the ORCCA baseline.}
    \setlength{\tabcolsep}{3.5pt}
\renewcommand{\arraystretch}{1.0}
\scriptsize
\begin{tabular}{l|rrrrrrr}
\toprule
 \textsc{stream}       &   ORCCA &   \textsc{no adapt} &   \textsc{no weight} &   \textsc{no split} &   \textsc{cascade} &   \textsc{kde} \\
\midrule
& \multicolumn{7}{c}{\textsc{known class balanced accuracy (criterion 1)}}  \\

 K2 N8 D0 &  0.974 &          -0.004 &                    +0.001 &     0.000 &                 -0.024 &         -0.097 \\
 K2 N8 D5 &  0.991 &          -0.218 &                    0.000 &     0.000 &                 -0.010 &         -0.042 \\ 
 K4 N6 D0 &  0.966 &          -0.006 &                    +0.002 &     0.000 &                 -0.067 &         -0.121 \\
 K4 N6 D5 &  0.986 &          -0.274 &                    +0.001 &     0.000 &                 -0.025 &         -0.152 \\
Covertype & 0.202  & +0.025          & +0.014                    & 0.000     & -0.010                 & -0.025         \\
\midrule
& \multicolumn{7}{c}{\textsc{known vs unknown class balanced accuracy (criterion 2)}}  \\

 K2 N8 D0 &  0.933 &          +0.005 &                    -0.010 &     0.000 &                 0.000 &         +0.001 \\
 K2 N8 D5 &  0.989 &          -0.149 &                    -0.001 &     0.000 &                 0.000 &         -0.014 \\ 
 K4 N6 D0 &  0.893 &          +0.007 &                    -0.013 &     0.000 &                 -0.001 &         +0.026 \\
 K4 N6 D5 &  0.980 &          -0.196 &                    -0.004 &     0.000 &                 +0.001 &         -0.053 \\
 Covertype & 0.723  & -0.045          & -0.014                    & 0.000     & 0.000                 & +0.020         \\
 \midrule
 & \multicolumn{7}{c}{\textsc{unknown class ARI (criterion 3)}}  \\
K2 N8 D0 &  0.443 &          -0.024 &                    -0.020 &     +0.002 &                 -0.008 &         -0.024 \\
 K2 N8 D5 &  0.483 &          -0.060 &                    -0.005 &     +0.006 &                 -0.009 &         -0.021 \\ 
 K4 N6 D0 &  0.378 &          -0.007 &                    -0.019 &     -0.003 &                 -0.007 &         +0.049 \\
 K4 N6 D5 &  0.480 &          -0.088 &                    +0.009 &     -0.001 &                 +0.006 &         +0.037 \\
Covertype & 0.162  & -0.050          & -0.043                    & -0.063     & -0.001                 & -0.089         \\

\bottomrule
\end{tabular}
    \label{tab:ablation}
\end{table}

The first column indicates the baseline score of \textsc{orcca}, and the following columns indicate the difference when a particular mechanism is switched off. The evaluated variants are (a) no adaptation ($p=0$), (b) constant fine-tuning probability without weighting, (c) no buffer splitting, i.e. assuming only unimodal novelties ($\beta = \infty$), (d) using cascade known class recognition instead of searching for minimal error, and (e) using only \textsc{kde} density with $\tau$ when recognizing both known and unknown classes. The last two versions aimed to validate the two-step recognition of \textsc{orcca}, where known classes are recognized using thresholds and unknowns -- with \textsc{kde}. In the (d) version, the first known model that accepts a sample will be used as a label. In (e), the recognition of knowns follows the same rule as unknowns. 

The values presented in the columns represent the difference between the baseline score and a given variant. Negatives mean that the version performs worse, and positives indicate an increase in quality. No adaptation (a) shows a significant loss in known class recognition and a known vs unknown dichotomy when the concept drift is introduced (streams with D5). This mechanism brings little difference when the concepts are static (streams with D0). Constant fine-tuning probability (b) only slightly lowers criterion~2, therefore, this mechanism may not be as crucial. Assuming a unimodal distribution (c) impacts only the \textsc{ari} criterion, since buffer-splitting only affects the recognition among novelties. As mentioned, synthetic streams show one novelty at a time, and the changes in their case are minor. In contrast, the result for Covertype, where multiple classes can co-occur at a time, validates the use of the buffer-splitting mechanism.  The last two variants (d) and (e) show a significant benefit of using the two-step mechanism, especially on criterion~1 and criterion~3 in the case of Covertype. This shows that treating labels provided in the offline phase to establish thresholds is less uncertain than relying on \textsc{kde} only. 

\section{Limitations}

The research using both real-world and a vast pool of synthetic and semi-synthetic data streams aimed to present a reliable view of the method. However, it is worth noting that the concept drifts and class distributions simulated on synthetic data streams may not match all real, open-world scenarios. The unsupervised adaptation to concept drift is possible due to its incremental nature, linked to a smooth distribution transition. In the case of gradual or sudden changes, instances of drifting classes could be recognized as novelties or anomalies. Validating adaptation to other types of concept drifts would require introducing the experimental protocol that involves feedback with partial supervision or validation of novelty occurrence.

Another limitation of the experimental protocol was linked to the way synthetic concept drift was simulated. The drift mechanism shifts classes in random directions, which usually increases their separation. Such a mechanism was intended to also reduce the possibility of overlapping the distributions of known classes with unknowns. The comparison with nonstationary concepts clearly rewards the adaptive mechanisms employed in \textsc{orcca}, \textsc{echo}, and \textsc{cnd}. Still, the use of synthetic and semi-synthetic data streams aimed to complement the real-world data stream experiments in a controlled environment, where the number of drifts and the number of novelties could be explicitly configured.

A particular limitation of \textsc{orcca} that became visible in the experiments is memory consumption. Even though the relative memory usage does not significantly increase compared to other methods, as presented in the experiments, the use of autoencoders requires the method to store the optimized neural network weights. The size of the state will grow with the increasing number of recognized classes. A mechanism for reducing memory usage, e.g. via knowledge distillation or aggregation of class-specific architectures, could reduce memory consumption. The mentioned limitations will be addressed in future research.

\section{Conclusions}

This work introduced, described, and evaluated \textit{Open-world Reconstruction-based Classifier with Concept Drift Adaptation} (\textsc{orcca}) method -- a classifier designed for novelty detection in an open-world data stream setting. The method uses an incrementally constructed ensemble of autoencoders and a two-step recognition mechanism that relies on thresholded reconstruction error to recognize known classes and error density established using \textsc{kde} to recognize unknowns and group them into separate classes. \textsc{Orcca} uses reconstruction error as a proxy instead of relying on the original features of samples. A particularly beneficial mechanism employed in the proposed approach is the adaptation to concept drifts, which uses adjusted probabilities for model fine-tuning based on the sample's reconstruction error. The method also uses Sarle's bimodality coefficient calculated on novel class errors to detect bimodal distributions, which are observed when more than one class appears at a given time.

The conducted studies used synthetic and semi-synthetic data streams based on Madelon distributions and \textsc{cifar-10} embeddings. The data stream generator simulated various numbers of known and unknown classes, as well as multiple concept drifts. Replicating the study allowed for stabilizing the results and performing statistical analysis. To support the observations from synthetic and semi-synthetic data, the research used three real-world data streams, where novel classes emerged. The experiments evaluated the proposed \textsc{orcca}, comparing it with seven reference methods in a three-criterion setting in terms of recognition quality, as well as time and memory usage. The results indicate that the proposed method achieves state-of-the-art results, especially when recognizing between known and novel instances, and shows time and memory growth indicators suitable for its effective application in a non-stationary streaming setting.

\section*{Acknowledgment}
This work was supported by the statutory funds of the Department of Systems and Computer Networks, Faculty of Information and Communication Technology, Wrocław University of Science and Technology.


\bibliographystyle{abbrv}
\bibliography{biblio}

@article{guyon2004result,
  title={Result analysis of the nips 2003 feature selection challenge},
  author={Guyon, Isabelle and Gunn, Steve and Ben-Hur, Asa and Dror, Gideon},
  journal={Advances in neural information processing systems},
  volume={17},
  year={2004}
}

@inproceedings{han2019learning,
  title={Learning to discover novel visual categories via deep transfer clustering},
  author={Han, Kai and Vedaldi, Andrea and Zisserman, Andrew},
  booktitle={2019 IEEE/CVF International Conference on Computer Vision (ICCV)},
  pages={8400--8408},
  year={2019},
  organization={IEEE}
}

@inproceedings{reiss2012introducing,
  title={Introducing a new benchmarked dataset for activity monitoring},
  author={Reiss, Attila and Stricker, Didier},
  booktitle={2012 16th international symposium on wearable computers},
  pages={108--109},
  year={2012},
  organization={IEEE}
}

@article{domingos2003general,
  title={A general framework for mining massive data streams},
  author={Domingos, Pedro and Hulten, Geoff},
  journal={Journal of Computational and Graphical Statistics},
  volume={12},
  number={4},
  pages={945--949},
  year={2003},
  publisher={Taylor \& Francis}
}

@inproceedings{tavallaee2009detailed,
  title={A detailed analysis of the {KDD CUP} 99 data set},
  author={Tavallaee, Mahbod and Bagheri, Ebrahim and Lu, Wei and Ghorbani, Ali A},
  booktitle={2009 IEEE symposium on computational intelligence for security and defense applications},
  pages={1--6},
  year={2009},
  organization={Ieee}
}

@article{blackard1999comparative,
  title={Comparative accuracies of artificial neural networks and discriminant analysis in predicting forest cover types from cartographic variables},
  author={Blackard, Jock A and Dean, Denis J},
  journal={Computers and electronics in agriculture},
  volume={24},
  number={3},
  pages={131--151},
  year={1999},
  publisher={Elsevier}
}

@inproceedings{komorniczak2025synthetic,
  title={Synthetic non-stationary data streams for recognition of the unknown},
  author={Komorniczak, Joanna},
  booktitle={Joint European Conference on Machine Learning and Knowledge Discovery in Databases},
  pages={143--159},
  year={2025},
  organization={Springer}
}

@article{montiel2021river,
  title={River: machine learning for streaming data in python},
  author={Montiel, Jacob and Halford, Max and Mastelini, Saulo Martiello and Bolmier, Geoffrey and Sourty, Raphael and Vaysse, Robin and Zouitine, Adil and Gomes, Heitor Murilo and Read, Jesse and Abdessalem, Talel and others},
  journal={Journal of Machine Learning Research},
  volume={22},
  number={110},
  pages={1--8},
  year={2021}
}

@inproceedings{barcina2024resilience,
  title={Resilience to the flowing unknown: An open set recognition framework for data streams},
  author={Barcina-Blanco, Marcos and L. Lobo, Jesus and Garcia-Bringas, Pablo and Del Ser, Javier},
  booktitle={International Conference on Hybrid Artificial Intelligence Systems},
  pages={139--152},
  year={2024},
  organization={Springer}
}

@article{ge2017generative,
  title={Generative openmax for multi-class open set classification},
  author={Ge, ZongYuan and Demyanov, Sergey and Chen, Zetao and Garnavi, Rahil},
  journal={arXiv preprint arXiv:1707.07418},
  year={2017}
}

@article{spinosa2009novelty,
  title={Novelty detection with application to data streams},
  author={Spinosa, Eduardo J and de Carvalho, Andr{\'e} Ponce de Leon F and Gama, Jo{\~a}o},
  journal={Intelligent Data Analysis},
  volume={13},
  number={3},
  pages={405--422},
  year={2009},
  publisher={SAGE Publications Sage UK: London, England}
}

@article{faria2016novelty,
  title={Novelty detection in data streams},
  author={Faria, Elaine R and Gon{\c{c}}alves, Isabel JCR and de Carvalho, Andr{\'e} CPLF and Gama, Jo{\~a}o},
  journal={Artificial Intelligence Review},
  volume={45},
  number={2},
  pages={235--269},
  year={2016},
  publisher={Springer}
}

@article{masud2010classification,
  title={Classification and novel class detection in concept-drifting data streams under time constraints},
  author={Masud, Mohammad and Gao, Jing and Khan, Latifur and Han, Jiawei and Thuraisingham, Bhavani M},
  journal={IEEE Transactions on knowledge and data engineering},
  volume={23},
  number={6},
  pages={859--874},
  year={2010},
  publisher={IEEE}
}

@article{cejnek2018concept,
  title={Concept drift robust adaptive novelty detection for data streams},
  author={Cejnek, Matous and Bukovsky, Ivo},
  journal={Neurocomputing},
  volume={309},
  pages={46--53},
  year={2018},
  publisher={Elsevier}
}

@article{shang2025novelty,
  title={Novelty-aware concept drift detection for neural networks},
  author={Shang, Dan and Zhang, Guangquan and Lu, Jie},
  journal={Neurocomputing},
  volume={617},
  pages={128933},
  year={2025},
  publisher={Elsevier}
}

@article{agrahari2022concept,
  title={Concept drift detection in data stream mining: A literature review},
  author={Agrahari, Supriya and Singh, Anil Kumar},
  journal={Journal of King Saud University-Computer and Information Sciences},
  volume={34},
  number={10},
  pages={9523--9540},
  year={2022},
  publisher={Elsevier}
}

@inproceedings{de2016iot,
  title={Iot big data stream mining},
  author={De Francisci Morales, Gianmarco and Bifet, Albert and Khan, Latifur and Gama, Joao and Fan, Wei},
  booktitle={Proceedings of the 22nd ACM SIGKDD international conference on knowledge discovery and data mining},
  pages={2119--2120},
  year={2016}
}

@article{ali2025novel,
  title={A novel framework for concept drift detection using autoencoders for classification problems in data streams},
  author={Ali, Usman and Mahmood, Tariq},
  journal={International Journal of Machine Learning and Cybernetics},
  volume={16},
  number={1},
  pages={397--418},
  year={2025},
  publisher={Springer}
}

@inproceedings{khan2009survey,
  title={A survey of recent trends in one class classification},
  author={Khan, Shehroz S and Madden, Michael G},
  booktitle={Irish conference on artificial intelligence and cognitive science},
  pages={188--197},
  year={2009},
  organization={Springer}
}

@article{sethi2015don,
  title={Don’t pay for validation: Detecting drifts from unlabeled data using margin density},
  author={Sethi, Tegjyot Singh and Kantardzic, Mehmed},
  journal={Procedia Computer Science},
  volume={53},
  pages={103--112},
  year={2015},
  publisher={Elsevier}
}

@inproceedings{oza2019c2ae,
  title={C2ae: Class conditioned auto-encoder for open-set recognition},
  author={Oza, Poojan and Patel, Vishal M},
  booktitle={Proceedings of the {IEEE/CVF} conference on computer vision and pattern recognition},
  pages={2307--2316},
  year={2019}
}

@inproceedings{gozuaccik2019unsupervised,
  title={Unsupervised concept drift detection with a discriminative classifier},
  author={G{\"o}z{\"u}a{\c{c}}{\i}k, {\"O}mer and B{\"u}y{\"u}k{\c{c}}ak{\i}r, Alican and Bonab, Hamed and Can, Fazli},
  booktitle={Proceedings of the 28th ACM international conference on information and knowledge management},
  pages={2365--2368},
  year={2019}
}

@article{scheirer2012toward,
  title={Toward open set recognition},
  author={Scheirer, Walter J and de Rezende Rocha, Anderson and Sapkota, Archana and Boult, Terrance E},
  journal={IEEE transactions on pattern analysis and machine intelligence},
  volume={35},
  number={7},
  pages={1757--1772},
  year={2012},
  publisher={IEEE}
}

@article{gomes2022survey,
  title={A survey on semi-supervised learning for delayed partially labelled data streams},
  author={Gomes, Heitor Murilo and Grzenda, Maciej and Mello, Rodrigo and Read, Jesse and Le Nguyen, Minh Huong and Bifet, Albert},
  journal={ACM Computing Surveys},
  volume={55},
  number={4},
  pages={1--42},
  year={2022},
  publisher={ACM New York, NY}
}

@article{gozuaccik2021concept,
  title={Concept learning using one-class classifiers for implicit drift detection in evolving data streams},
  author={G{\"o}z{\"u}a{\c{c}}{\i}k, {\"O}mer and Can, Fazli},
  journal={Artificial Intelligence Review},
  volume={54},
  number={5},
  pages={3725--3747},
  year={2021},
  publisher={Springer}
}

@inproceedings{bifet2007learning,
  title={Learning from time-changing data with adaptive windowing},
  author={Bifet, Albert and Gavalda, Ricard},
  booktitle={Proceedings of the 2007 SIAM international conference on data mining},
  pages={443--448},
  year={2007},
  organization={SIAM}
}

@inproceedings{gama2004learning,
  title={Learning with drift detection},
  author={Gama, Joao and Medas, Pedro and Castillo, Gladys and Rodrigues, Pedro},
  booktitle={Brazilian symposium on artificial intelligence},
  pages={286--295},
  year={2004},
  organization={Springer}
}

@inproceedings{menon2020concept,
  title={Concept drift detection in phishing using autoencoders},
  author={Menon, Aditya Gopal and Gressel, Gilad},
  booktitle={Symposium on Machine Learning and Metaheuristics Algorithms, and Applications},
  pages={208--220},
  year={2020},
  organization={Springer}
}

@inproceedings{jaworski2020concept,
  title={Concept drift detection using autoencoders in data streams processing},
  author={Jaworski, Maciej and Rutkowski, Leszek and Angelov, Plamen},
  booktitle={International Conference on Artificial Intelligence and Soft Computing},
  pages={124--133},
  year={2020},
  organization={Springer}
}

@inproceedings{khadimallah2024autoencoder,
  title={Autoencoder-Based Drift Detection Method for Dynamic Analysis of {EEG} Data: A Comprehensive Study},
  author={Khadimallah, Rihab and Kallel, Ilhem and Medina, Javier Sanchez and Drira, Fadoua},
  booktitle={2024 IEEE International Conference on Systems, Man, and Cybernetics (SMC)},
  pages={4654--4659},
  year={2024},
  organization={IEEE}
}

@article{lukats2024,
	title = {A benchmark and survey of fully unsupervised concept drift detectors on real-world data streams},
	issn = {2364-4168},
	doi = {10.1007/s41060-024-00620-y},
	journal = {International Journal of Data Science and Analytics},
	author = {Lukats, Daniel and Zielinski, Oliver and Hahn, Axel and Stahl, Frederic},
	month = aug,
	year = {2024},
}

@article{gaudreault2024systematic,
  title={A systematic literature review of novelty detection in data streams: Challenges and opportunities},
  author={Gaudreault, Jean-Gabriel and Branco, Paula},
  journal={ACM Computing Surveys},
  volume={56},
  number={10},
  pages={1--37},
  year={2024},
  publisher={ACM New York, NY}
}

@article{brzezinski2019dynamics,
  title={On the dynamics of classification measures for imbalanced and streaming data},
  author={Brzezinski, Dariusz and Stefanowski, Jerzy and Susmaga, Robert and Szczech, Izabela},
  journal={IEEE transactions on neural networks and learning systems},
  volume={31},
  number={8},
  pages={2868--2878},
  year={2019},
  publisher={IEEE}
}

@inproceedings{cao2006density,
  title={Density-based clustering over an evolving data stream with noise},
  author={Cao, Feng and Ester, Martin and Qian, Weining and Zhou, Aoying},
  booktitle={Proceedings of the 2006 SIAM international conference on data mining},
  pages={328--339},
  year={2006},
  organization={SIAM}
}

@inproceedings{aggarwal2003framework,
  title={A framework for clustering evolving data streams},
  author={Aggarwal, Charu C and Philip, S Yu and Han, Jiawei and Wang, Jianyong},
  booktitle={Proceedings 2003 VLDB conference},
  pages={81--92},
  year={2003},
  organization={Elsevier}
}

@inproceedings{haque2016efficient,
  title={Efficient handling of concept drift and concept evolution over stream data},
  author={Haque, Ahsanul and Khan, Latifur and Baron, Michael and Thuraisingham, Bhavani and Aggarwal, Charu},
  booktitle={2016 IEEE 32nd international conference on data engineering (ICDE)},
  pages={481--492},
  year={2016},
  organization={IEEE}
}

@article{de2016minas,
  title={{MINAS}: multiclass learning algorithm for novelty detection in data streams},
  author={de Faria, Elaine Ribeiro and Ponce de Leon Ferreira Carvalho, Andr{\'e} Carlos and Gama, Joao},
  journal={Data mining and knowledge discovery},
  volume={30},
  number={3},
  pages={640--680},
  year={2016},
  publisher={Springer}
}

@article{demvsar2006statistical,
  title={Statistical comparisons of classifiers over multiple data sets},
  author={Dem{\v{s}}ar, Janez},
  journal={Journal of Machine learning research},
  volume={7},
  number={Jan},
  pages={1--30},
  year={2006}
}

\end{document}